\documentclass{article}

\usepackage{arxiv}

\usepackage[utf8]{inputenc} % allow utf-8 input
\usepackage[T1]{fontenc}    % use 8-bit T1 fonts
\usepackage{hyperref}       % hyperlinks
\usepackage{url}            % simple URL typesetting
\usepackage{booktabs}       % professional-quality tables
\usepackage{amsfonts}       % blackboard math symbols
\usepackage{nicefrac}       % compact symbols for 1/2, etc.
\usepackage{microtype}      % microtypography
\usepackage{lipsum}
\usepackage{amsmath,amssymb}
\usepackage[dvipsnames]{xcolor}
\usepackage{subfiles} 
\usepackage{pgfplots}
\pgfplotsset{width=90mm, compat=1.9}
\usepackage{pifont}
\usepackage{url}

\title{Poly-YOLO: higher speed, more precise detection and instance segmentation for YOLOv3}

\author{
  Petr Hurtik*, Vojtech Molek*, Jan Hula*, Marek Vajgl*, Pavel Vlasanek\thanks{University of Ostrava, Centre of Excellence IT4Innovations, Institute for Research and Applications of Fuzzy Modeling, 30. dubna 22, Ostrava, Czech Republic}, and Tomas Nejezchleba\thanks{Varroc Lighting Systems, Suvorovova 195, Šenov u Nového Jičína, Czech Republic.}
}

\begin{document}
\maketitle

\begin{abstract}
We present a new version of YOLO with better performance and extended with instance segmentation called Poly-YOLO. Poly-YOLO builds on the original ideas of YOLOv3 and removes two of its weaknesses: a large amount of rewritten labels and inefficient distribution of anchors. Poly-YOLO reduces the issues by aggregating features from a light SE-Darknet-53 backbone with a hypercolumn technique, using stairstep upsampling, and produces a single scale output with high resolution. In comparison with YOLOv3, Poly-YOLO has only 60\% of its trainable parameters but improves mAP by a relative 40\%. We also present Poly-YOLO lite with fewer parameters and a lower output resolution. It has the same precision as YOLOv3, but it is three times smaller and twice as fast, thus suitable for embedded devices. Finally, Poly-YOLO performs instance segmentation using bounding polygons. The network is trained to detect size-independent polygons defined on a polar grid. Vertices of each polygon are being predicted with their confidence, and therefore Poly-YOLO produces polygons with a varying number of vertices. Source code is available at \url{https://gitlab.com/irafm-ai/poly-yolo}.
\end{abstract}

% keywords can be removed

\begin{figure}[h]
    \centering
    \includegraphics[width=165mm]{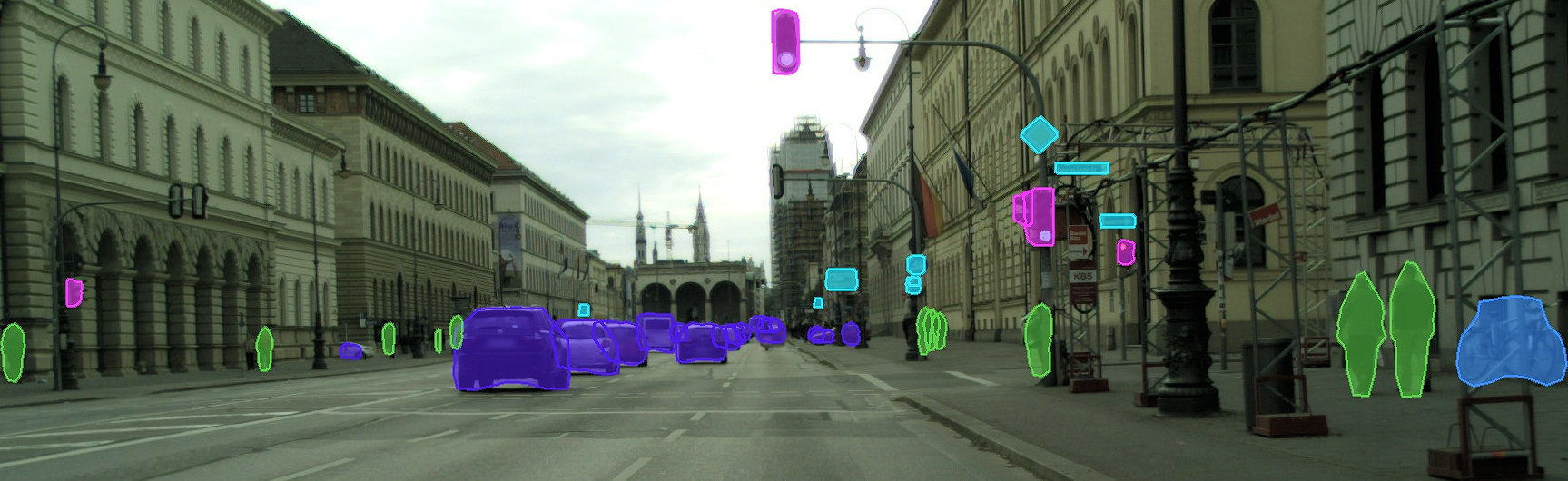}
\caption{The figure shows instance segmentation performance of the proposed Poly-YOLO algorithm applied on Cityscapes dataset and running 22FPS on a mid-tier graphic card. Image was cropped due to visibility.}
\label{fig-intro}
\end{figure}

\section{Problem statement}
\label{sec-problem-statement}

Object detection is a process where all important areas containing objects of interest are bounded while the background is ignored. Usually, the object is bounded by a box that is expressed in terms of spatial coordinates of its top-left corner and its width and height. The disadvantage of this approach is that for the objects of complex shapes, the bounding box also includes background, which can occupy a significant part of the area as the bounding box does not wrap the object tightly. Such behavior can decrease the performance of a classifier applied over the bounding box~\cite{hurtik2020yoloasc} or may not fulfill requirements of precise detection~\cite{kirillov2019panoptic}. To avoid the problem, classical detectors such as Faster R-CNN~\cite{ren2015faster} or RetinaNet~\cite{lin2017focal} were modified into a version of Mask R-CNN~\cite{he2017mask} or RetinaMask~\cite{fu2019retinamask}. These methods also infer the instance segmentation, i.e., each pixel in the bounding box is classified into object/background classes.
The limitation of the methods is their computation speed, where they are unable to reach real-time performance on non-high-tier hardware. The problem we focus on is to create a precise detector with instance segmentation and the ability of real-time processing on mid-tier graphic cards.

In this study, we start with YOLOv3~\cite{redmon2018yolov3}, which excels in processing speed, and therefore it is a good candidate for real-time applications running on computers~\cite{huang2018yolo} or mobile devices~\cite{oltean2019towards}. On the other hand, the precision of YOLOv3 lags behind detectors such as RetinaNet~\cite{lin2017focal}, EfficientDet~\cite{tan2019efficientdet}, or CornerNet~\cite{law2018cornernet}. We analyze YOLO's performance and identify its two drawbacks. The first drawback is low precision of the detection of big boxes~\cite{redmon2018yolov3} caused by inappropriate handling of anchors in output layers. The second one is rewriting of labels by each-other due to the coarse resolution. To solve these issues, we design a new approach, dubbed Poly-YOLO, that significantly pushes forward original YOLOv3 abilities. To tackle the problem of instance segmentation, we propose a way to detect tight polygon-based contour. Our contributions and benefits of our approach are as follows:

\begin{itemize}
\item we propose Poly-YOLO that increases the detection accuracy of the previous version, YOLOv3. Poly-YOLO has a brand-new feature decoder with a single output tensor that goes to head with higher resolution that solves two principal YOLO's issues: rewriting of labels and incorrect distribution of anchors. 
\item We produce a single output tensor by a hypercolumn composition of multi-resolution feature maps produced by a feature extractor. To unify the resolutions of the feature maps, we utilize stairstep upscaling, which allows us to obtain slightly lower loss with comparison to direct upscaling while the computation speed is preserved.
\item We design an extension that realizes instance segmentation using bounding polygon representation. The number of maximal polygon vertices can be adjusted according to a requirement to a precision. 
\item The bounding polygon is detected within a polar grid with relative coordinates that allow the network to learn general, size-independent shapes. The network produces a dynamic number of vertices per bounding polygon.
\end{itemize}

\begin{figure}[!th]
    \centering
    \includegraphics[width=54mm]{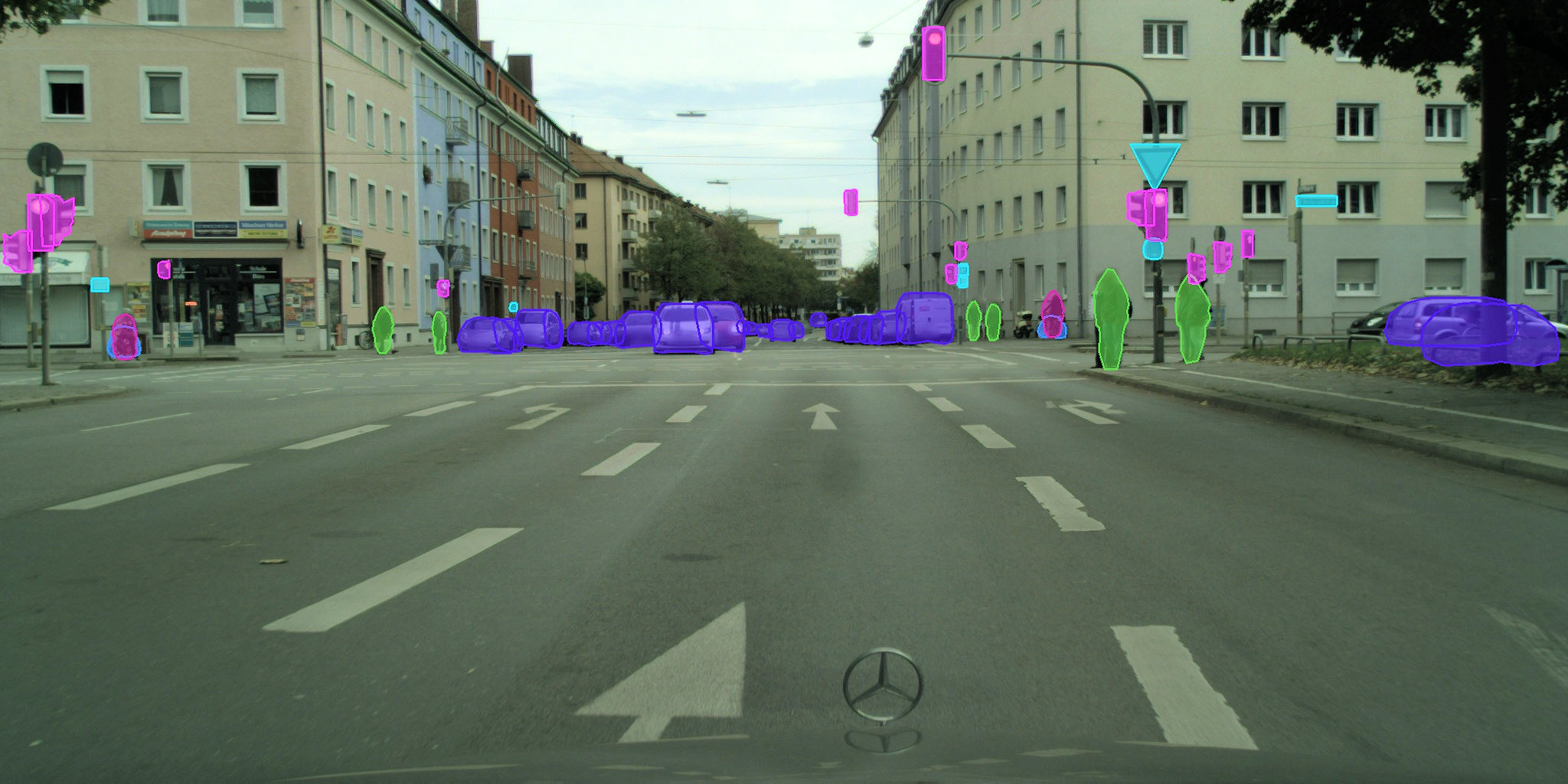}
    \includegraphics[width=54mm]{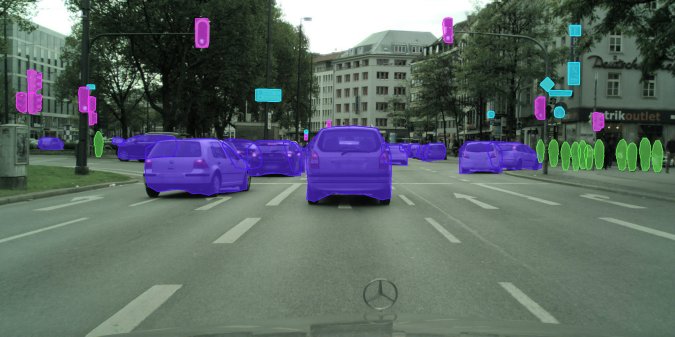}
    \includegraphics[width=54mm]{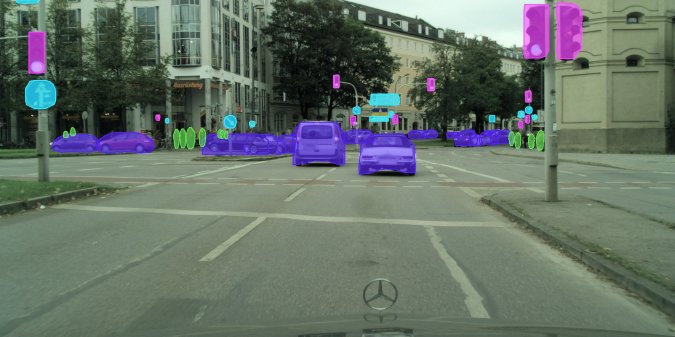}
    \caption{Examples of Poly-YOLO inference on the Cityscapes testing dataset.}
    \vspace{5mm}
    \includegraphics[width=54mm]{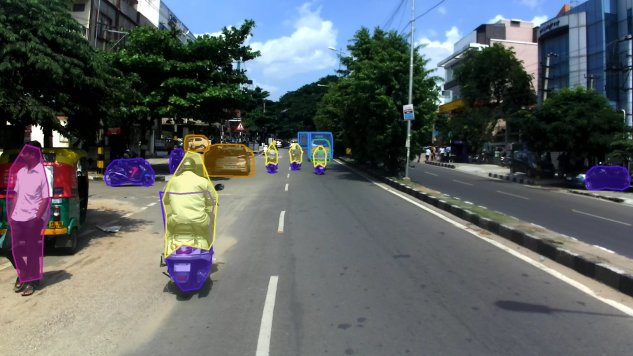}
    \includegraphics[width=54mm]{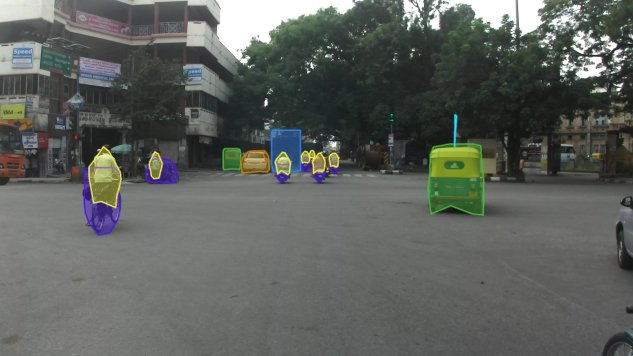}
    \includegraphics[width=54mm]{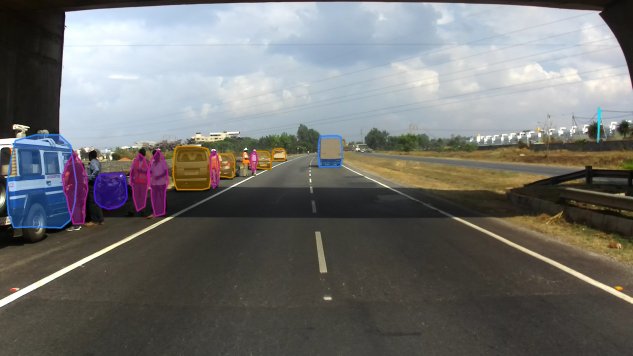}
    \caption{Examples of Poly-YOLO inference on the India driving testing dataset.}
    %\mv{obrázk zvážit cuty, na těchnle jsou ty detekovaé objekty zbytečně malé a není to moc vidět}
    \label{fig-demonstration}
\end{figure}

\section{Current state and related work}

\subsection{Object detection}
Models for object detection can be divided into two groups, two-stage, and one-stage detectors. Two-stage detectors split the process as follows. In the first phase, regions of interest (RoI) are proposed, and in the subsequent stage, bounding box regression and classification is being done inside these proposed regions. One-stage detectors predict the bounding boxes and their classes at once. Two-stage detectors are usually more precise in terms of localization and classification accuracy, but in terms of processing are slower then one-stage detectors. Both of these types contain a backbone network for feature extraction and head networks for classification and regression. Typically, the backbone is some SOTA network such as ResNet~\cite{he2017mask} or ResNext~\cite{xie2017aggregated}, pre-trained on ImageNet or OpenImages. Even though, some approaches~\cite{shen2019object}, \cite{zhu2019scratchdet} also experiment with training from scratch.

\subsubsection{Two-stage detectors}
The prototypical example of two-stage architecture is Faster R-CNN~\cite{ren2015faster}, which is an improvement of its predecessor Fast R-CNN~\cite{girshick2015fast}. The main improvement lies in the use of Region Proposal Network (RPN), which replaced a much slower selective search of RoIs. It also introduced the usage of multi-scale anchors to detect objects of different sizes. %It is trained using a multi-task loss function, which optimizes four losses at once. Two losses are computed for the proposed RoIs by the RPN where the first is for RoI regression, and the second is for object/background classification of the ROI. Another two losses are computed for the final classification score and final box coordinates. 
Faster R-CNN is, in a way, a meta-algorithm that can have many different incarnations depending on a type of the backbone and its heads. One of the frequently used backbones, called Feature Pyramid Network (FPN)~\cite{dollar2014fast}, allows to predict RoIs from multiple feature maps, each with a different resolution. This is beneficial for the recognition of objects at different scales. 

\subsubsection{One-stage detectors}
Two best-known examples of one-stage detectors are YOLO~\cite{redmon2018yolov3} and SSD~\cite{liu2016ssd}. The architecture of YOLO will be thoroughly described in Section~\ref{sec-yolo}. Usually, one-stage detectors divide the image into a grid and predict bounding boxes and their classes inside them, all at once. Most of them also use the concept of anchors, which are predefined typical dimensions of bounding boxes that serve as apriori knowledge. One of the major improvements in the area of one-stage detectors was a novel loss function call Focal Loss \cite{lin2017focal}. Because of the fact that two-stage detectors produce a sparse set of region proposals in the first step, most of the negative locations are filtered out for the second stage. One-stage detectors, on the other hand, produce a dense set of region proposals which they need to classify as containing objects or not. This creates a problem with the non-proportional frequency of negative examples. Focal Loss solves this problem by adjusting the importance of negative and positive examples within the loss function. 
Another interesting idea was proposed in an architecture called RefineDet~\cite{zhang2018single}, which performs a two-step regression of the bounding boxes. The second step refines the bounding boxes proposed in the first step, which produces more accurate detection, especially for small objects.
Recently, there has been a surge of interest in approaches that do not use anchor boxes. The main representative of this trend is the FCOS framework~\cite{tian2019fcos}, which works by predicting four coordinates of a bounding box for every foreground pixel. These four coordinates represent a distance to the four boundary edges of a bounding box in which the pixel is enclosed in. The predicted bounding boxes of every pixel are subsequently filtered by NMS. Similar anchor-free approach was proposed in CornerNet~\cite{law2018cornernet}, where the objects are detected as a pair of top-left and bottom-right corners of a bounding box.

%GraphNets, NAS, EfficientDet
%Combination of 1 and 2 stage detectors
%SlowFast Networks for Video Recognition

%Pedestrian detection: An evaluation of the state of the art %\cite{dollar2011pedestrian}

%Faster R-CNN \cite{ren2015faster}

%Object detection with discriminatively trained part-based models %\cite{felzenszwalb2009object}

%Fast feature pyramids for object detection \cite{dollar2014fast}

%Object detection in videos by high quality object linking \cite{tang2019object}

%What Makes for Effective Detection Proposals? \cite{hosang2015makes}

%Region-Based Convolutional Networks for Accurate Object Detection and Segmentation \cite{girshick2015region}

%Mask R-CNN \cite{he2017mask}

%Object Detection Networks on Convolutional Feature Maps \cite{ren2017objectdet}

\subsection{Instance Segmentation}
In many applications, a boundary given by a rectangle may be too crude, and we may instead require a boundary framing the object tightly. In the literature, this task is called Instance Segmentation, and the main approaches also fit into the one-stage/two-stage taxonomy. The prototypical example of a two-stage method is an architecture called Mask R-CNN~\cite{he2017mask}, which extended Faster R-CNN by adding a separate fully-convolutional head that predicts masks of objects. Note, the same principle is also applied to RetinaNet, and the improved net is called RetinaMask~\cite{fu2019retinamask}. One of Mask R-CNN innovations is a novel way for extracting features from RoIs using the RoIAlign layer, which avoids the problem of misalignments of the RoI due to its quantization to the grid of the feature map. 
One-stage methods for instance segmentation can be further divided into top-down methods, bottom-up methods, and direct methods. Top-down methods~\cite{chen2019tensormask,bolya2019yolact} work by first detecting an object and then segmenting this object within a bounding box. Prediction of bounding boxes either uses anchors or is anchor free following the FCOS framework~\cite{tian2019fcos}. Bottom-up methods~\cite{newell2017associative,liu2017sgn}, on the other hand, work by first embedding each pixel into a metric space in which are these pixels subsequently clustered. As the name suggests, direct methods work by directly predicting the segmentation mask without bounding boxes or pixel embedding~\cite{wang2020solov2}. We also mention that independently of our instance segmentation, PolarMask~\cite{xie2019polarmask} introduces instance segmentation using polygons, which are also predicted in polar coordinates. In comparison with PolarMask, Poly-YOLO learns itself in general size-independent shapes due to the use of the relative size of a bounding polygon according to the particular bounding box. The second difference is that Poly-YOLO produces a dynamic number of vertices per polygon, according to the shape-complexity of various objects.

\section{Fast and precise object detection with Poly-YOLO}
\label{sec-yolo}
Here, we firstly recall YOLOv3 fundamental ideas, describe issues that block reaching higher performance, and propose our solution that removes them.

\subsection{YOLO history}
\begin{figure}[ht]
    \centering
    \includegraphics[width=.48\textwidth]{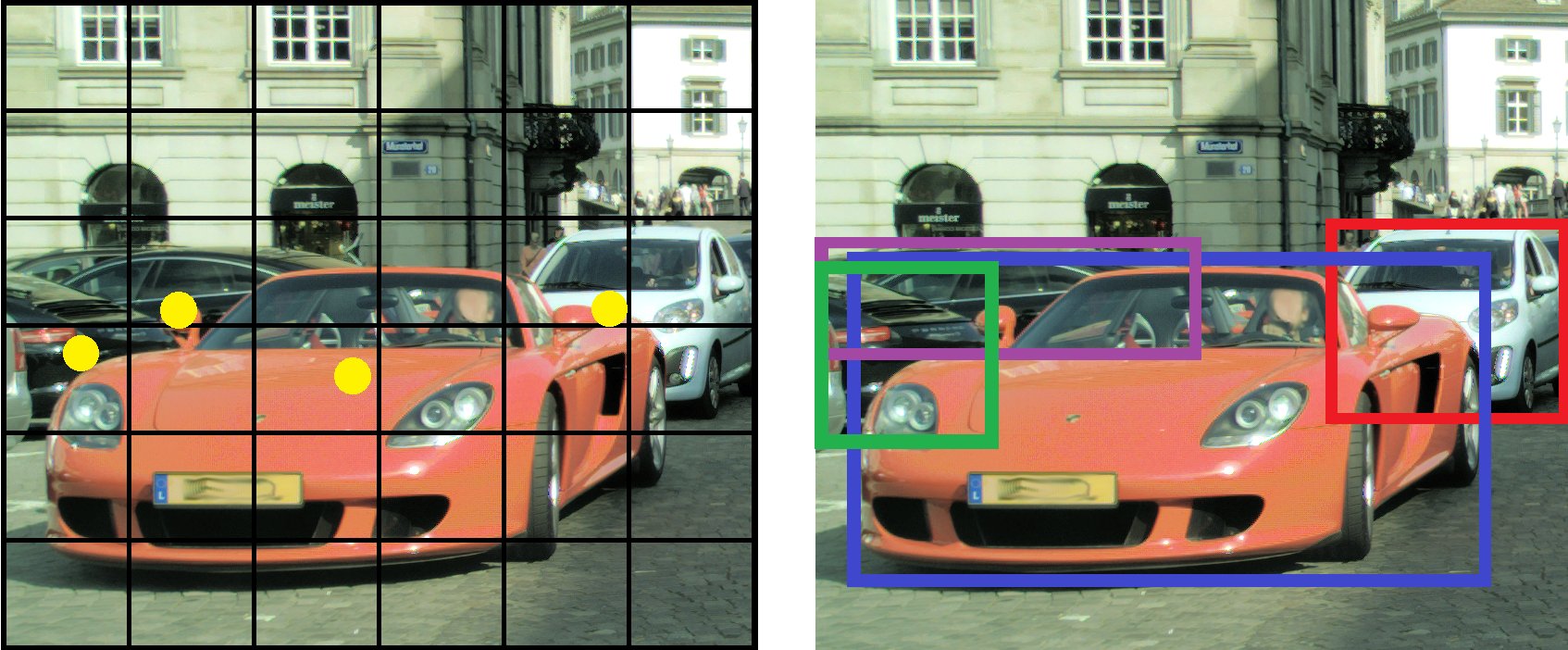}
    \caption{The left image illustrates the YOLO grid over the input image, and yellow dots represent centers of detected objects. The right image illustrates detections.}
    \label{fig:yolov1^grid}
\end{figure}

First version of YOLO (You Only Look Once) was introduced in 2016~\cite{redmon2016you}. The motivation behind YOLO is to create a fast object detector with an emphasis on speed. The detector is made of two essential parts: the convolutional neural network (CNN) and specially designed loss function. The CNN backbone is inspired by GoogleNet~\cite{szegedy2015going} and has 24 convolutional layers followed by 2 fully connected layers. The network output is reshaped into two dimensional \textit{grid} with with the shape of $G^h \times G^w$, where $G^h$ is number of cells in vertical side and $G^w$ in horizontal side. Each grid cell occupies a part of the image, as depicted in Fig.~\ref{fig:yolov1^grid}.
Every object in the image has its center in one of the cells, and that particular cell is responsible for detecting and classifying said object. More precisely, the responsible cell outputs $N^B$ bounding boxes. Each box is given as a tuple $(x,y,w,h)$ and a confidence measure. Here, $(x,y)$ is the center of the predicted box relative to the cell boundary and $(w,h)$ is the width and height of the bounding box relative to the image size. The confidence measures how much is the cell confident that it contains an object. Finally, each cell outputs $N^c$ conditional class probabilities, i.e. probabilities that detected object belongs to certain class(es). 
In other words, cell confidence tells us that there is object in the predicted box and conditional class probabilities tells us that the box contains, e.g., vehicle -- car. 
The final output of the model is a tensor with dimensions $G^h \times G^w \times (5N^B+N^c)$, where constant five is used because of $(x,y,w,h)$ and a confidence.% All the above-described network output semantics is embedded in the loss function.

YOLOv2~\cite{redmon2017yolo9000} brought a couple of improvements. Firstly, the architecture of the convolutional neural network was updated to Darknet-19 -- a fully convolutional network with 19 convolutional layers containing batch normalization and five max-pooling layers. The cells are no longer predicting plain $(x,y,w,h)$ directly, but instead scales and translates \textit{anchor boxes}. The parameters $(a^w, a^h)$, i.e., width and height of an anchor box for all anchors boxes are extracted from a training dataset with usage of $k$-means algorithm. The clustering criterion is IoU. %Now each cell predicts for each assigned anchor box . %Box prediction is then given as $$(\sigma(t^x) + c^x, \sigma(t^y) + c^y, p^we_{t^w}, p^he_{t^h}, \sigma(t^o))$$ where $\sigma(\cdot)$ is logistic function, $\sigma(t^o)$ is a confidence, $(c^x,c^y)$ is distance of cell from grid origin and $(p^w, p^h)$ is width and height of anchor box. 
Lastly, YOLOv2 uses skip connections to concatenate features from different parts of the CNN to create final tensor of feature maps, including features across different scales and levels of abstraction.

The most recent version of YOLO~\cite{redmon2018yolov3} introduces mainly three output scales and a deeper architecture -- Darknet-53. Each of the scale/feature-map has its own set of anchors -- three per output scale. Compared with v2, YOLOv3 reaches higher accuracy, but due to the heavier backbone, its inference speed is decreased.

\subsection{YOLOv3 issues blocking better performance}
\label{subsec-yolov3-issues}
YOLOv3, as it is designed, suffers from two issues that we discovered and that are not described in the original papers: rewriting of labels and imbalanced distribution of anchors across output scales. Solving these issues is crucial for improvement of the YOLO performance.

\begin{figure}[t]
    \centering
    \includegraphics[width=88mm]{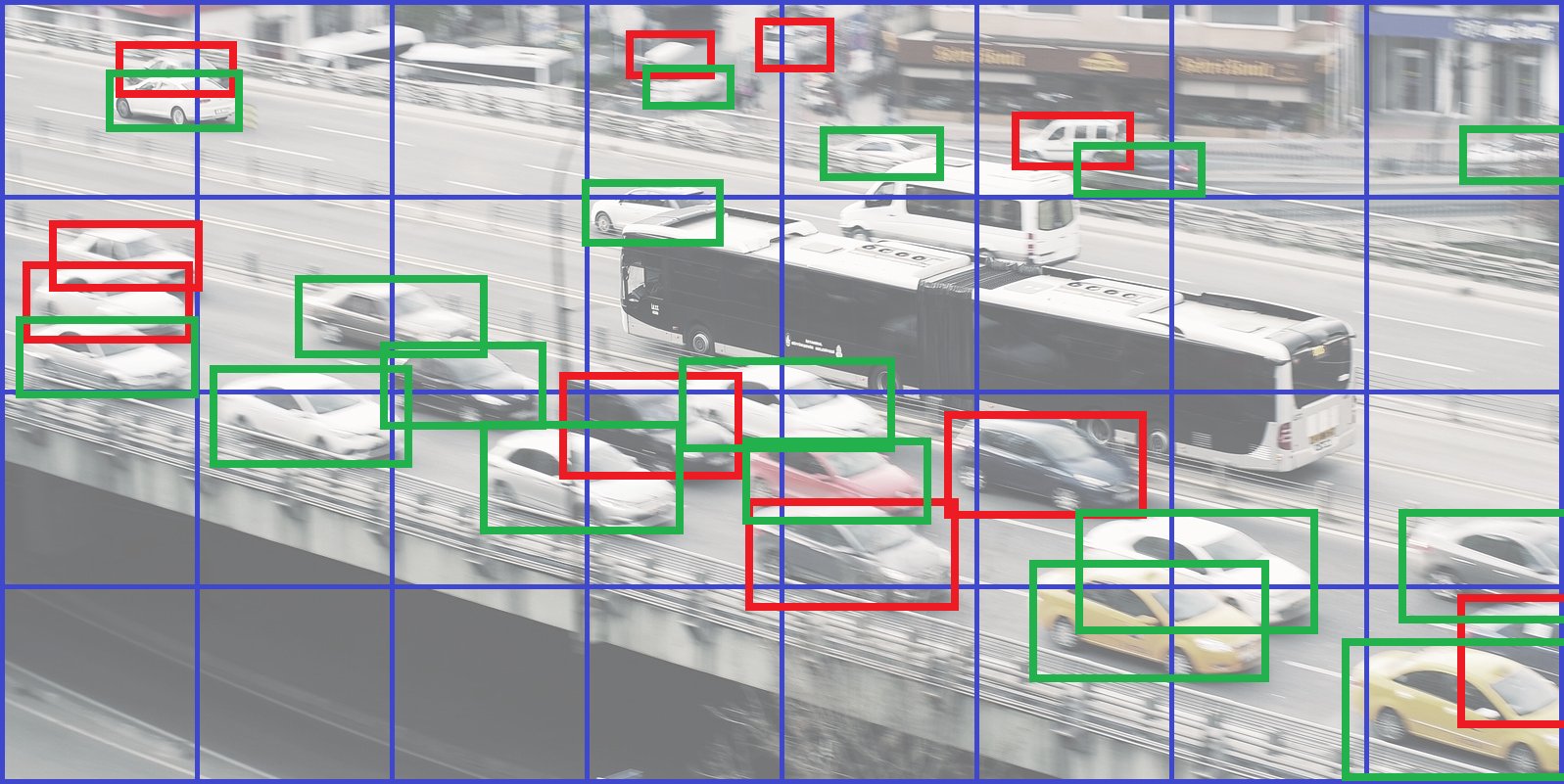}
\caption{The image illustrates the label rewriting problem for the detection of cars. A label is rewritten by other if centers of two boxes (with the same anchor box) belong to the same cell. In this illustrative example, blue denotes grid, red rewritten label, and green preserved label. Here, 10 labels out of 27 are rewritten, and the detector is not trained to detect them.}
\label{fig-overwrite}
\end{figure}

\subsubsection{Label rewriting problem} here, we discuss situation, when a bounding box given by its label from a ground truth dataset can be rewritten by other box and therefore the network is not trained to detect it. For the sake of simplicity and explanation, we avoid the usage of the anchors notation in the text bellow. Let us suppose an input image with a resolution of $r\times r$ pixels. Furthermore, let $s_k$ be a scale ratio of an $k$-th output to the input, where YOLOv3 uses the following ratios:  $s_1=1/8, s_2=1/16, s_3=1/32$. These scales are given by the YOLOv3 architecture, namely by strided convolutions. Finally, let $B=\{\mathbf{b}_1, \dots, \mathbf{b}_n\}$ be a set of boxes presented in an image. Each box $\mathbf{b}_i$ is represented as a tuple $(b_i^{x^1}, b_i^{y^1}, b_i^{x^2}, b_i^{y^2})$ that defines its top-left and bottom-right corner. For simplicity, we also derive centers $C=\{\mathbf{c}_1, \dots, \mathbf{c}_n\}$ where $\mathbf{c}_i = (c_i^x, c_i^y)$ is defined as $c_i^x =  0.5(b_i^{x^1}+b_i^{x^2})$ and the same for $c_i^y$. With this notation, label is rewritten, if the following holds:
\begin{equation}
\exists(\mathbf{c}_i, \mathbf{c}_j \in C): \xi(c_i^x, c_j^x, s_k) + \xi(c_i^y, c_j^y, s_k)=2,
\label{lab-rewrite}
\end{equation}
where
\begin{equation}
\xi(x, y, z) = 
\begin{cases}
    1, &  \lfloor xz \rfloor = \lfloor yz \rfloor\\
    0, & \text{else}
\end{cases},
\label{lab-xi-function}
\end{equation}
and $\lfloor \cdot \rfloor$ denotes the lowest integer of the term.
The purpose of function $\xi$ is to check if both boxes are assigned to the same cell of a grid on the scale $s_k$.  In simple words, if two boxes on the same scale are assigned to the same cell, then one of them will be rewritten. Introducing anchors, both must belong to the same anchor.
As a consequence, the network is trained to ignore some objects, which leads to a low number of positive detections. According to Equations~\eqref{lab-rewrite} and \eqref{lab-xi-function}, there is a crucial role of $s_k$ that directly affects the number and the resolution of cells. Considering standard resolution of YOLO $r=416$, then, for $s_3$ (the coarsest scale) we obtain a grid of $13\times13$ cells with size of $32\times32$ pixels each. Also, the absolute size of boxes does not affect the label rewriting problem; the important indicator is the box center. The practical illustration for such a setting and its consequence for the labels is shown in Figure~\ref{fig-overwrite}. The ratio of rewritten labels in the datasets used in the benchmark is shown in Table~\ref{table-rewrite}.

\begin{table}
\begin{center}
{\caption{Amount of rewritten labels for various datasets}
\label{table-rewrite}}
\renewcommand{\arraystretch}{1.05}
\begin{tabular}{ll|rrr}
%\hline
&&\multicolumn{3}{c}{Rewritten labels [\%]}\\
%\cline{2-4}
&&&Poly&Poly\\
Dataset &Resolution&YOLOv3&YOLO&YOLO lite\\
\cline{1-5}
Simulator &416$\times$416  &16.36 &0.22&2.31\\
Simulator &608$\times$800  &12.55 &0.00&0.61\\
\cline{1-5}
Cityscapes &416$\times$416 &9.51 &2.79&9.50\\
Cityscapes &608$\times$832 &3.92 &0.97&2.75\\
Cityscapes &640$\times$1280&2.56 &0.59&1.44\\
\cline{1-5}
India Driving& 416$\times$416  &23.07&5.80&13.78\\
India Driving & 448$\times$800 &13.54 &1.92&~~4.96\\
India Driving & 704$\times$1280&9.16&1.12&~~2.44\\
\end{tabular}
\end{center}
\end{table}

\subsubsection{Anchors distribution problem}
the second YOLO issue comes from the fact that YOLO is anchor-based (i.e., it needs prototypical anchor boxes for training/detection), and the anchors are distributed among output scales. Namely, YOLOv3 uses nine anchors, three per output scale. A particular ground truth box is matched with the best matching anchor that assigns it to a certain output layer scale. Here, let us suppose a set of boxes sizes $M=\{\mathbf{m}_1,\dots,\mathbf{m}_n\}$, where $\mathbf{m}_i=(m_i^w, m_i^h)$ is given by $m_i^w = b_i^{x^2}-b_i^{x^1}$ for width and analogously for height.  The $k$-means algorithm~\cite{macqueen1967some} is applied to $M$ in order to determine centroids in 2D space, which then represent the nine anchors. The anchors are split into triplets and connected with small, medium, and large boxes detected in the output layers. Unfortunately, such a principle of splitting anchors according to three sizes is generally reasonable if
$$
M\sim\mathcal{U}(0,r)
$$
holds. By $\mathcal{U}(0,r)$ we notate a uniform distribution between bounds given by 0 and $r$. But, such the condition cannot be guaranteed for various applications in general. Note, $M~\sim \mathcal{N}(0.5r,r)$, where $\mathcal{N}(0.5r,r)$ is normal distribution with mean $\mu = 0.5r$ and standard deviation $\sigma^2 = r$ is a more realistic case, which causes that most of the boxes will be captured by the middle output layer (for the medium size) and the two other layers will be underused.

\begin{figure*}[!ht]
    \includegraphics[width=165mm]{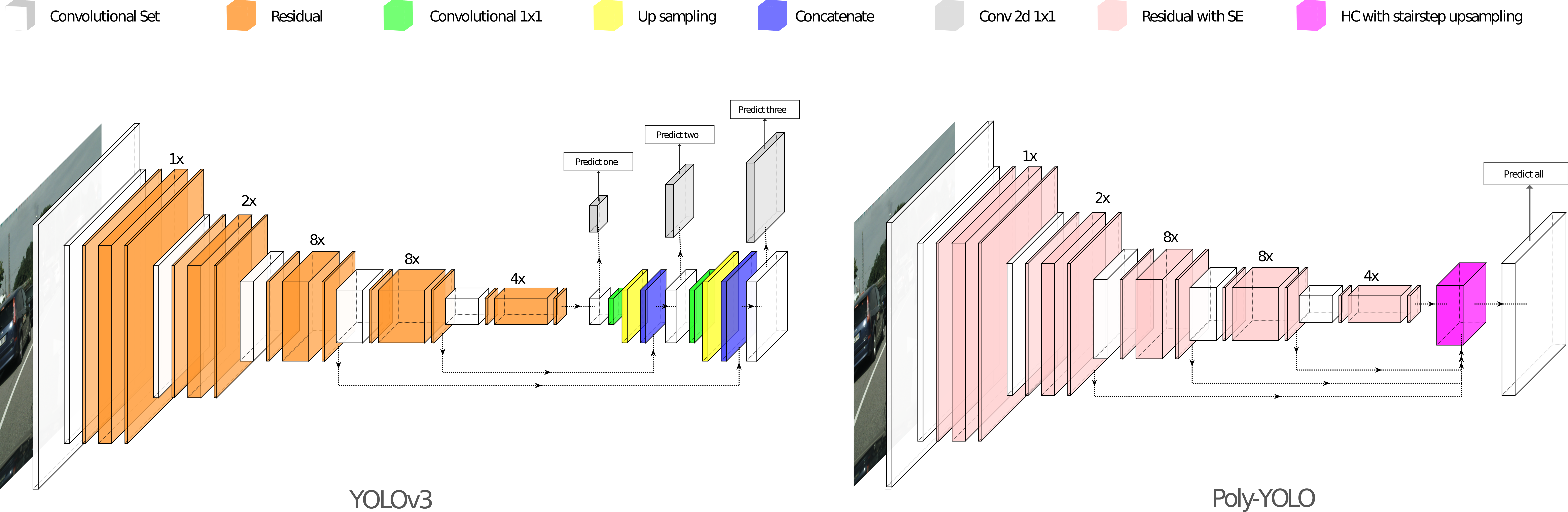}
\caption{A comparison of YOLOv3 and Poly-YOLO architecture. Poly-YOLO uses less convolutional filters per layer in the feature extractor part and extends it by squeeze-and-excitation blocks.  The heavy neck is replaced by a lightweight block with hypercolumn that utilizes a stairstep for upsampling. The head now uses single instead of three outputs and has a higher resolution. In summary, Poly-YOLO has 40\% less parameters than YOLOv3 while producing more precise predictions.}
\label{fig-architecture^new}
\end{figure*}

To illustrate the problem, let us suppose two sets of box sizes, $M_1$ and $M_2$; the former connected with the task of car plate detection from a camera placed over the highway and the latter connected with a person detection from a camera placed in front of the door. For such tasks, we can obtain roughly $M_1~\sim \mathcal{N}(0.3r,0.2r)$ because the plates will cover small areas and $M_2~\sim \mathcal{N}(0.7r,0.2r)$ because the people will cover large areas. For both sets, anchors are computed separately. The first case leads to the problem that output scales for medium and large will also include small anchors because the dataset does not include big objects. Here, the problem of label rewriting will escalate because small objects will need to be detected in a coarse grid. The second case works vice-versa. Large objects will be detected in small and medium output layers. Here, detection will not be precise because small and medium output layers have limited receptive fields. The receptive field for the three used scales is \{$85\times85$, $181\times181$, $365\times365$\}. The practical impact of the two cases is the same: performance will be sub-optimal. In the paper that introduced YOLOv3~\cite{redmon2018yolov3}, the author says \emph{"YOLOv3 has relatively high AP$_{\text{small}}$ performance. However, it has comparatively worse performance on medium and larger size objects. More investigation is needed to get to the bottom of this"}. We believe that the reason why YOLOv3 has these problems is explained in the paragraph above.

\subsection{Poly-YOLO architecture}
Before we describe the architecture itself, let us mention the motivation and the justification for it. As we described in the previous section, YOLO's performance suffers from the problem of label rewriting and the problematic distribution of anchors among output scales. 

The first issue can be suppressed by high values of $s_k$, i.e., a scale multiplicator that expresses the ratio of output resolution with respect to the input resolution $r$. The ideal case would happen when $r=rs_k$, i.e., $s_k=1$, which means that output and input resolutions are equal. In this case, no label rewriting may occur. Such a condition is generally held in many encoder-decoder-based segmentation NNs such as U-Net~\cite{ronneberger2015u}. As we are focusing on computational speed, we have to omit such a scheme to find a solution where $s_k<1$ will be a reasonable trade-off. Let us recall that YOLOv3 uses $s_1=1/8$, $s_2=1/16$, $s_3=1/32$.

The second issue can be solved using one of two ways. The first way is to define receptive fields for the three output scales, and define two thresholds that will split them. Then, $k$-means will compute centroids triplets (used as anchors) according to these thresholds. This would change the data-driven anchors to problem-driven (receptive field) anchors.  For example, data $M~\sim \mathcal{N}(r/5,r/10)$  would be detected only on scale detecting small objects and not on all scales as it is currently realized in YOLOv3. The drawback of such a way is that we will not use a full capacity of the network. The second way is to create an architecture with a single output that will aggregate information from various scales. Such an aggregated output will also handle all the anchors at once. So, in contrast to the first way, the estimation of anchor sizes will be again data-driven.

We propose to use a single output layer with a high $\mathbf{s_1}$ scale ratio connected with all the anchors, which solves both issues mentioned above. Namely, we use $s_1=1/4$. An illustration of a comparison between the original and the new architecture is shown in Figure~\ref{fig-architecture^new}. For the composition of the single output scale from multiple partial scales, we use the hypercolumn technique~\cite{hariharan2016object}. Formally, let $O$ be a feature map, $u(\cdot, \omega)$ function upscaling an input by a factor $\omega$, and $m(\cdot)$ be a function transforming feature map with dimensions $a \times b \times c \times \cdot$ into a feature map with dimensions $a \times b \times c \times \delta$, where $\delta$ is a constant. Furthermore, we consider $g(O_1,\dots, O_n)$ to be an $n$-nary composition/aggregation function. For that, the output feature map using the hypercolumn is given as
\begin{equation*}
O = g\left(m\left(O_1\right), u\left(m(O_2), 2^1\right), \dots, u\left(m(O_n), 2^{n-1}\right)\right).
\end{equation*}
Selecting addition as an aggregation function, the formula can be rewritten as
\begin{equation*}
O = \sum_{i=1}^{n} u(m(O_{i}),2^{i-1}).
\end{equation*}

\begin{figure}[!h]
    \centering
    \includegraphics[width=140mm]{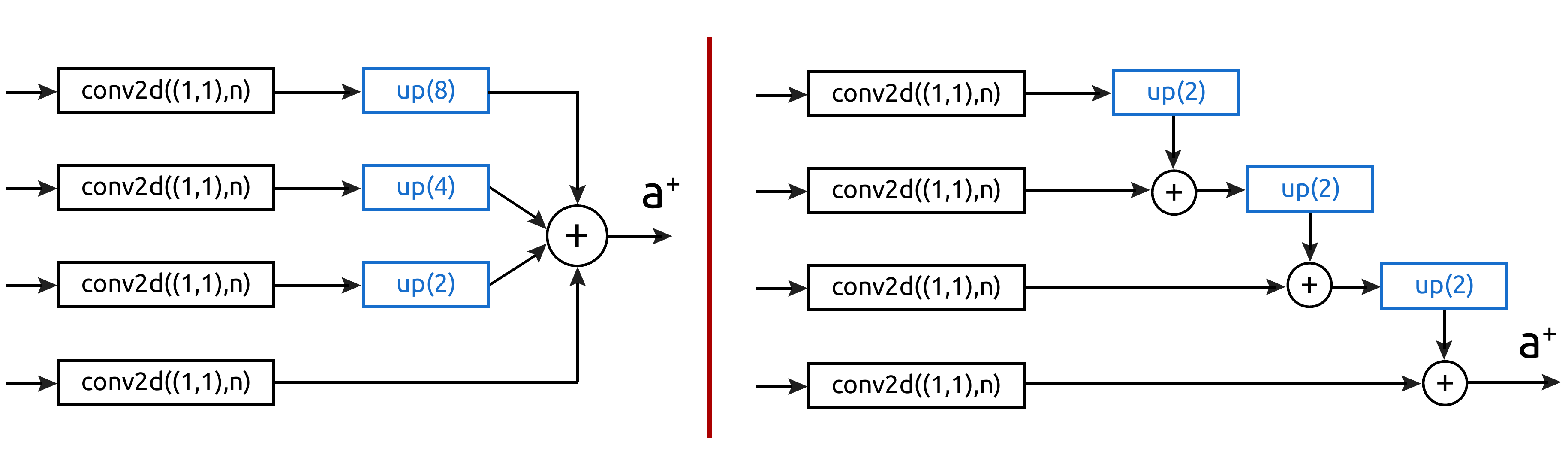}
\caption{Illustration of HC scheme (left) and HC with stairstep (right).}
\label{fig-hc-vs-hc-stairstep}
\end{figure}

As it is evident from the formula, there is a high imbalance - a single value of $O_1$ projects into $O$ just single value, while a single value of  $O_n$ is projected into $2_{n-1}\times2_{n-1}$ values directly. To break the imbalance, we propose to use the staircase approach known from the computer graphic, see Figure~\ref{fig-hc-vs-hc-stairstep}. The stairstep interpolation increases (or decreases for downscale) an image resolution by 10\% at maximum until the desired resolution is reached. In comparison with a direct upscale, the output is more smooth but does not include,e.g., step artifacts as a direct upsampling does. Here, we will use the lowest available upscale factor, two. Formally, stairstep output feature map $O'$ is defined as
\begin{gather*}
O'= \dots u(u(m(O_{n}), 2) + m(O_{n-1}),2)\dots + m(O_1).
\end{gather*}
If we consider the nearest neighbor upsampling, $O=O'$ holds. For bilinear interpolation (and others), $O\neq O'$ is reached for non-homogenous inputs. The critical fact is that the computational complexity is equal for both direct upscaling and stairstep upscaling. Although the stairstep approach realizes more adding, they are computed over feature maps with a lower resolution, so the number of added elements is identical.

\begin{figure}[!h]
\centering
\begin{tikzpicture}
\small
\begin{axis}[height=70mm,width=120mm,xlabel={Epoch},xmin=15, xmax=50, ymin=6.7,ymax=10.0,legend style={at={(0.79,1.05)},anchor=north, nodes={scale=0.9, transform shape}},legend columns=1, ymajorgrids=true,yminorgrids=true,style={ultra thick},axis x line*=bottom, axis y line*=left, legend cell align={left}]
\addplot[purple]coordinates{(0,0)};\addlegendentry{HC loss}
\addplot[violet]coordinates{(0,0)};\addlegendentry{HC val loss}
\addplot[green]coordinates{(0,0)};\addlegendentry{HC StairStep loss}
\addplot[OliveGreen]coordinates{(0,0)};\addlegendentry{HC StairStep val loss}
\addplot[purple,line width=0.01mm, opacity=0.4]coordinates{(1,2.732029413604736305e+02)(2,2.813758148193359432e+01)(3,1.865099098205566364e+01)(4,1.658868531227111731e+01)(5,1.532994195938110416e+01)(6,1.439157340049743716e+01)(7,1.374231960296630817e+01)(8,1.278546295166015589e+01)(9,1.204399978637695234e+01)(10,1.148155005455017097e+01)(11,1.101463806152343672e+01)(12,1.052351519584655826e+01)(13,1.016023430824279750e+01)(14,9.824525899887085245e+00)(15,9.664322261810303516e+00)(16,9.384550552368164489e+00)(17,9.239576935768127441e+00)(18,9.005624904632568928e+00)(19,8.588596863746642995e+00)(20,8.571057310104370330e+00)(21,8.404228963851927858e+00)(22,8.340165009498596405e+00)(23,8.186875495910644673e+00)(24,8.208669204711913636e+00)(25,8.208902883529663796e+00)(26,8.077020092010497976e+00)(27,7.989252104759215989e+00)(28,7.941696276664734100e+00)(29,7.888717341423034490e+00)(30,7.832836790084838796e+00)(31,7.832175683975219904e+00)(32,7.663851766586303960e+00)(33,7.559942717552185165e+00)(34,7.547161898612976039e+00)(35,7.519452729225158727e+00)(36,7.540841507911681951e+00)(37,7.467275552749633505e+00)(38,7.462745251655579004e+00)(39,7.435551800727844274e+00)(40,7.317458891868591664e+00)(41,7.177728977203368999e+00)(42,7.175092720985412953e+00)(43,7.266593136787414586e+00)(44,7.227136011123657333e+00)(45,7.138212966918945135e+00)(46,7.057641859054565536e+00)(47,7.091027755737304261e+00)(48,7.075931344032287740e+00)(49,7.004830484390258682e+00)(50,7.019328255653380921e+00)};
\addplot[violet,line width=0.01mm, opacity=0.4]coordinates{(1,4.967645057678222997e+01)(2,2.051643737792968736e+01)(3,2.244594039916992045e+01)(4,5.953843406677246008e+01)(5,1.561156095504760799e+01)(6,1.381811599731445384e+01)(7,1.317733169555664041e+01)(8,1.277578512191772475e+01)(9,1.228445928573608370e+01)(10,1.123451930999755888e+01)(11,1.086640621185302713e+01)(12,1.067012592315673913e+01)(13,1.042972589492797830e+01)(14,1.025514892578124915e+01)(15,9.696580963134765696e+00)(16,9.795581703186035938e+00)(17,1.040123163223266545e+01)(18,1.002047077178955092e+01)(19,9.243096523284911825e+00)(20,9.162713146209716442e+00)(21,9.154740371704100710e+00)(22,8.802516250610350923e+00)(23,9.044807281494140483e+00)(24,8.755937767028807883e+00)(25,8.674948072433471324e+00)(26,8.561986923217773438e+00)(27,8.904793062210082155e+00)(28,8.404041852951049663e+00)(29,8.800080013275145774e+00)(30,8.596466121673584482e+00)(31,8.561056137084960938e+00)(32,8.525397510528565093e+00)(33,8.433293657302856516e+00)(34,8.271793947219848064e+00)(35,8.530273542404174236e+00)(36,8.225887775421142578e+00)(37,8.426810569763183878e+00)(38,8.425880107879638814e+00)(39,8.251130285263061026e+00)(40,8.284497613906859570e+00)(41,8.055769214630126740e+00)(42,8.453122224807739116e+00)(43,8.301476068496704386e+00)(44,8.144826278686522869e+00)(45,8.126517381668090678e+00)(46,8.063105964660644887e+00)(47,8.120507183074950674e+00)(48,7.961165361404418661e+00)(49,8.143256092071533558e+00)(50,7.954543809890747497e+00)};
\addplot[green, line width=0.01mm, opacity=0.4]coordinates{(1,2.452838519287109307e+02)(2,2.755705022811889648e+01)(3,1.861096101760864130e+01)(4,1.666367929458618136e+01)(5,1.548004856109619176e+01)(6,1.455340915679931690e+01)(7,1.353859471321105978e+01)(8,1.263150207519531243e+01)(9,1.195589118957519581e+01)(10,1.134866663932800357e+01)(11,1.080311893463134787e+01)(12,1.043550397872924762e+01)(13,9.987273645401000266e+00)(14,9.722882604598998668e+00)(15,9.516404981613158398e+00)(16,9.199944734573364258e+00)(17,9.076932945251463991e+00)(18,8.904954447746277424e+00)(19,8.746439037322998544e+00)(20,8.598318157196045419e+00)(21,8.560267624855042001e+00)(22,8.417530617713929075e+00)(23,8.353507542610168812e+00)(24,8.116536803245544718e+00)(25,8.240563883781433674e+00)(26,7.827806768417358008e+00)(27,7.768239860534667862e+00)(28,7.780174336433410609e+00)(29,7.709556579589843750e+00)(30,7.609791646003722931e+00)(31,7.574410209655761506e+00)(32,7.624542207717895259e+00)(33,7.587248201370239364e+00)(34,7.352278175354004297e+00)(35,7.261359257698059011e+00)(36,7.346882891654968439e+00)(37,7.261687479019165004e+00)(38,7.303116822242737172e+00)(39,7.127460465431213699e+00)(40,7.038791823387145818e+00)(41,7.068486633300781641e+00)(42,7.022750735282897949e+00)(43,6.985782737731933700e+00)(44,6.979401111602783203e+00)(45,6.966746344566344895e+00)(46,6.989722766876220739e+00)(47,6.959112300872802415e+00)(48,6.925235023498535369e+00)(49,6.914559588432312331e+00)(50,6.887992649078369389e+00)};
\addplot[OliveGreen, line width=0.01mm, opacity=0.4]coordinates{(1,4.542122688293456889e+01)(2,3.208812026977538778e+01)(3,1.846792335510253835e+01)(4,1.649663707733154183e+01)(5,1.517077556610107436e+01)(6,1.485974163055419872e+01)(7,1.323084390640258867e+01)(8,1.252315515518188427e+01)(9,1.275436105728149450e+01)(10,1.173643318176269545e+01)(11,1.075469245910644567e+01)(12,1.045949918746948271e+01)(13,1.004454458236694414e+01)(14,1.148925222396850643e+01)(15,1.029657762527465792e+01)(16,9.812391510009765838e+00)(17,1.041357362747192461e+01)(18,9.272860803604125124e+00)(19,1.002135105133056570e+01)(20,8.920948753356933381e+00)(21,9.697752704620361541e+00)(22,8.797823352813720632e+00)(23,8.800837011337280913e+00)(24,9.014848976135253267e+00)(25,9.362278575897216015e+00)(26,8.407340612411498881e+00)(27,8.582284431457519958e+00)(28,8.454475021362304332e+00)(29,8.405545682907105132e+00)(30,8.191084661483763796e+00)(31,8.479650983810424592e+00)(32,8.421799621582032103e+00)(33,8.356268291473389453e+00)(34,8.103541984558106037e+00)(35,7.964511594772338832e+00)(36,8.060608539581298970e+00)(37,8.085789356231689595e+00)(38,8.069555740356445739e+00)(39,8.098917274475097017e+00)(40,8.104276962280273722e+00)(41,7.984401969909668217e+00)(42,7.900894880294799805e+00)(43,7.928395500183105682e+00)(44,7.826800575256347869e+00)(45,7.937250776290893484e+00)(46,7.941140527725219833e+00)(47,7.940442008972167898e+00)(48,7.836724977493286559e+00)(49,7.794039421081542685e+00)(50,7.766075038909912465e+00)};
\addplot[purple,line width=0.01mm, opacity=0.4]coordinates{(1,2.528245626068115257e+02)(2,2.743733217239379840e+01)(3,1.834135786056518569e+01)(4,1.636268758773803711e+01)(5,1.529552467346191413e+01)(6,1.413989991188049267e+01)(7,1.346058795928955121e+01)(8,1.289499230384826589e+01)(9,1.220603596687316816e+01)(10,1.156302303314208935e+01)(11,1.113940753936767614e+01)(12,1.075599851608276403e+01)(13,1.025146638870239180e+01)(14,9.941507406234741495e+00)(15,9.708590774536132173e+00)(16,9.378386545181275125e+00)(17,9.207190322875977273e+00)(18,9.059863309860229919e+00)(19,8.988651900291442232e+00)(20,8.766833143234253001e+00)(21,8.578268370628357786e+00)(22,8.495649461746216602e+00)(23,8.418717832565308257e+00)(24,8.486407127380370952e+00)(25,8.283108048439025595e+00)(26,8.251531982421875355e+00)(27,8.174219288825987917e+00)(28,8.164903979301453063e+00)(29,8.084267444610595987e+00)(30,7.892197051048278844e+00)(31,8.002829570770263246e+00)(32,7.810328249931335343e+00)(33,7.945059032440185298e+00)(34,7.955346064567565989e+00)(35,7.811710796356201136e+00)(36,7.479792065620422647e+00)(37,7.441696763038635254e+00)(38,7.490529184341430557e+00)(39,7.441144647598266992e+00)(40,7.425695953369140234e+00)(41,7.393730545043945668e+00)(42,7.213589811325073597e+00)(43,7.141757478713989471e+00)(44,7.112771601676940669e+00)(45,7.128325424194335902e+00)(46,7.169629101753234757e+00)(47,7.147744288444519256e+00)(48,7.082879333496093999e+00)(49,6.963916773796081650e+00)(50,6.901088352203369247e+00)};
\addplot[violet,line width=0.01mm, opacity=0.4]coordinates{(1,4.607471000671386463e+01)(2,2.084548404693603629e+01)(3,2.051848117828369311e+01)(4,1.649981880187988281e+01)(5,1.508177288055419929e+01)(6,1.526411630630493121e+01)(7,1.329909893035888757e+01)(8,1.257086112976074155e+01)(9,1.233927644729614315e+01)(10,1.146779685974121143e+01)(11,1.155708549499511761e+01)(12,1.050842901229858484e+01)(13,1.049309192657470646e+01)(14,1.009810781478881836e+01)(15,9.792327423095702699e+00)(16,9.731249980926513388e+00)(17,9.466507759094238139e+00)(18,9.569999980926514382e+00)(19,9.389018411636351757e+00)(20,9.250175590515135937e+00)(21,8.912770318984986062e+00)(22,9.470625896453856996e+00)(23,9.511586132049560049e+00)(24,8.773455438613892099e+00)(25,8.711553897857665874e+00)(26,8.863530502319335369e+00)(27,8.752632303237914968e+00)(28,8.697214946746825603e+00)(29,8.674681701660155397e+00)(30,8.647565145492553285e+00)(31,8.777959127426147035e+00)(32,8.411166687011718324e+00)(33,8.554255037307738618e+00)(34,8.725251598358154226e+00)(35,9.154226665496826953e+00)(36,8.234573488235474414e+00)(37,8.807540016174316477e+00)(38,8.206633758544921164e+00)(39,8.420194702148437571e+00)(40,8.578857078552246662e+00)(41,8.452195682525633913e+00)(42,8.208015413284302397e+00)(43,7.902823181152343501e+00)(44,7.924762268066405824e+00)(45,7.960425958633423171e+00)(46,7.849107561111449805e+00)(47,7.915323686599731623e+00)(48,8.015262746810913441e+00)(49,7.983601303100585689e+00)(50,7.734088630676269283e+00)};
\addplot[green,line width=0.01mm, opacity=0.4]coordinates{(1,2.714134547042846748e+02)(2,2.864620761871337962e+01)(3,1.831226913452148608e+01)(4,1.647569742202758647e+01)(5,1.528221658706664954e+01)(6,1.411544672966003411e+01)(7,1.327134907722473223e+01)(8,1.255360980033874441e+01)(9,1.178260114669799741e+01)(10,1.115518136024475027e+01)(11,1.061720126152038546e+01)(12,1.029654678344726548e+01)(13,9.949484148025511843e+00)(14,9.599913864135741548e+00)(15,9.477884292602539062e+00)(16,9.237170066833495952e+00)(17,8.971624770164488893e+00)(18,8.902279596328735778e+00)(19,8.723038654327393004e+00)(20,8.726411185264588255e+00)(21,8.519875769615174121e+00)(22,8.142673978805541779e+00)(23,8.007665672302245241e+00)(24,7.988226337432861257e+00)(25,7.940012784004211177e+00)(26,7.900626373291015803e+00)(27,7.624943008422851776e+00)(28,7.610255584716797195e+00)(29,7.570171184539795206e+00)(30,7.518637704849242986e+00)(31,7.506844835281372141e+00)(32,7.564294209480285858e+00)(33,7.457772641181946049e+00)(34,7.426055798530578400e+00)(35,7.482665400505066344e+00)(36,7.276001329421997177e+00)(37,7.236599888801574565e+00)(38,7.188371334075927876e+00)(39,7.176593265533447230e+00)(40,7.112188410758972346e+00)(41,7.122617630958557378e+00)(42,7.159048571586608567e+00)(43,7.058293752670287802e+00)(44,7.097299861907958807e+00)(45,7.070134911537170552e+00)(46,7.059775700569153045e+00)(47,7.081138987541198304e+00)(48,6.995380992889404403e+00)(49,7.021183810234069789e+00)(50,7.002889685630798056e+00)};
\addplot[OliveGreen,line width=0.01mm, opacity=0.4]coordinates{(1,5.182226661682128821e+01)(2,1.949191219329834013e+01)(3,1.809748615264892635e+01)(4,1.621714448928833008e+01)(5,1.601471281051635742e+01)(6,1.498474685668945305e+01)(7,1.326904109954833899e+01)(8,1.246662109374999972e+01)(9,1.214078098297119190e+01)(10,1.109432956695556705e+01)(11,1.109475891113281243e+01)(12,1.039105165481567461e+01)(13,1.012115404129028384e+01)(14,1.008565971374511783e+01)(15,9.728734130859374218e+00)(16,9.235529022216796946e+00)(17,9.523416728973389311e+00)(18,9.172139949798584269e+00)(19,9.979730377197265412e+00)(20,9.792681865692138388e+00)(21,9.179889812469482990e+00)(22,8.777746715545653444e+00)(23,8.423356904983521076e+00)(24,9.082673597335816140e+00)(25,8.772354555130004172e+00)(26,8.582319669723510458e+00)(27,8.435451536178588228e+00)(28,8.449435329437255504e+00)(29,8.281421213150023775e+00)(30,8.277170391082764311e+00)(31,8.464032621383667632e+00)(32,8.023574199676513530e+00)(33,8.297125129699706392e+00)(34,8.244062938690184694e+00)(35,8.114789171218871999e+00)(36,8.309085474014281658e+00)(37,8.235922441482543377e+00)(38,8.065134773254394318e+00)(39,7.943887214660644958e+00)(40,8.020460824966431090e+00)(41,7.939175033569336293e+00)(42,7.949748029708862518e+00)(43,8.037183399200438672e+00)(44,7.894220466613769638e+00)(45,7.941810722351074325e+00)(46,7.891608476638793945e+00)(47,7.827375917434692632e+00)(48,7.984894428253173970e+00)(49,7.849083452224731694e+00)(50,7.965461015701293945e+00)};
\addplot[purple,line width=0.01mm, opacity=0.4]coordinates{(1,2.351914439773559593e+02)(2,2.684756097793579244e+01)(3,1.834960435867309414e+01)(4,1.647380594253539954e+01)(5,1.522322324752807532e+01)(6,1.426686566352844210e+01)(7,1.320563168525695730e+01)(8,1.242509613990783635e+01)(9,1.173491564750671401e+01)(10,1.124532270431518555e+01)(11,1.078555891990661664e+01)(12,1.022917422294616685e+01)(13,1.001330440521240206e+01)(14,9.703085184097290039e+00)(15,9.531065645217895010e+00)(16,9.187894115447997834e+00)(17,9.001517133712768626e+00)(18,8.815730791091919372e+00)(19,8.731871418952941610e+00)(20,8.555791697502137083e+00)(21,8.412219448089599538e+00)(22,8.387536091804504323e+00)(23,8.292408881187439462e+00)(24,8.208884978294372914e+00)(25,8.221968812942504812e+00)(26,8.095802912712096955e+00)(27,8.062361397743224245e+00)(28,7.727567763328551997e+00)(29,7.703861742019653569e+00)(30,7.660030083656311106e+00)(31,7.559545407295226838e+00)(32,7.491743712425232182e+00)(33,7.507783226966857626e+00)(34,7.542719798088073802e+00)(35,7.511546430587768164e+00)(36,7.258367595672607031e+00)(37,7.204462313652038929e+00)(38,7.182908825874328329e+00)(39,7.155122442245483327e+00)(40,7.155898442268371795e+00)(41,7.102123932838440012e+00)(42,7.075755567550658931e+00)(43,7.009285402297973810e+00)(44,6.981239447593688752e+00)(45,6.976480593681335662e+00)(46,6.913790163993835058e+00)(47,6.915366749763489196e+00)(48,6.924115056991577255e+00)(49,6.918503575325011923e+00)(50,6.847625684738159357e+00)};
\addplot[violet,line width=0.01mm, opacity=0.4]coordinates{(1,4.145711982727051037e+01)(2,2.052312484741210952e+01)(3,2.563174240112304858e+01)(4,1.566165918350219677e+01)(5,1.929599464416503807e+01)(6,1.402692119598388665e+01)(7,1.438735919952392628e+01)(8,1.222263759613037060e+01)(9,1.173313674926757777e+01)(10,1.202093362808227539e+01)(11,1.086642854690551729e+01)(12,1.038208711624145586e+01)(13,1.030128892898559556e+01)(14,9.870856552124022798e+00)(15,9.612740535736083558e+00)(16,9.575021400451660725e+00)(17,9.499965553283692188e+00)(18,9.541812553405762287e+00)(19,9.282160301208495667e+00)(20,9.257449855804443573e+00)(21,9.295666007995604829e+00)(22,8.923839330673217773e+00)(23,8.986859292984009429e+00)(24,8.627354536056518697e+00)(25,8.633327417373656942e+00)(26,8.909267034530639506e+00)(27,8.686467113494872549e+00)(28,8.274678115844727344e+00)(29,8.352428169250488921e+00)(30,8.203520803451537446e+00)(31,8.226595230102539347e+00)(32,8.071482248306274698e+00)(33,8.371796436309814737e+00)(34,8.642152004241943075e+00)(35,8.124961700439452983e+00)(36,8.007710304260253764e+00)(37,8.394551687240600302e+00)(38,8.016853017807006765e+00)(39,7.909426345825194993e+00)(40,8.017199649810791584e+00)(41,8.204456558227539276e+00)(42,8.029864721298217489e+00)(43,7.862603254318237589e+00)(44,7.957250852584838441e+00)(45,7.995654458999633896e+00)(46,7.777288246154784979e+00)(47,7.780826139450073065e+00)(48,7.749791679382323828e+00)(49,7.897765769958495774e+00)(50,7.904670534133910742e+00)};
\addplot[green,line width=0.01mm, opacity=0.4]coordinates{(1,2.956282328796386878e+02)(2,2.905912553787231545e+01)(3,1.838769777297973462e+01)(4,1.634959197998047031e+01)(5,1.504833388328552246e+01)(6,1.393326461791992266e+01)(7,1.285141297340393152e+01)(8,1.209887529373168924e+01)(9,1.144876506805419858e+01)(10,1.089240140914916921e+01)(11,1.041564451217651310e+01)(12,1.003431389808654828e+01)(13,9.721194248199463317e+00)(14,9.453299608230590678e+00)(15,9.197321052551268750e+00)(16,9.017037310600279909e+00)(17,8.905480394363403107e+00)(18,8.692533254623413086e+00)(19,8.650345478057861470e+00)(20,8.588412055969238068e+00)(21,8.386691370010375124e+00)(22,8.188864517211914773e+00)(23,8.243088512420653657e+00)(24,8.112474513053893332e+00)(25,8.117725858688354279e+00)(26,8.070703105926513743e+00)(27,8.074334959983826110e+00)(28,8.077845897674560405e+00)(29,7.641695661544799734e+00)(30,7.516779441833495667e+00)(31,7.551589117050171218e+00)(32,7.598917145729064515e+00)(33,7.442061476707458212e+00)(34,7.408707818984985316e+00)(35,7.466960120201110662e+00)(36,7.405881485939025843e+00)(37,7.311071429252624654e+00)(38,7.212791976928710547e+00)(39,7.051171803474426625e+00)(40,7.077417092323303294e+00)(41,7.094711389541625834e+00)(42,7.076410803794860982e+00)(43,7.099491472244262802e+00)(44,7.038612179756164267e+00)(45,6.979040484428406188e+00)(46,6.980861501693725657e+00)(47,6.937830524444580149e+00)(48,6.868106570243835307e+00)(49,6.813048720359802246e+00)(50,6.794434719085693075e+00)};
\addplot[OliveGreen,line width=0.01mm, opacity=0.4]coordinates{(1,4.979463378906250171e+01)(2,2.008909221649170007e+01)(3,1.991720558166504063e+01)(4,1.555554735183715742e+01)(5,1.438377830505371158e+01)(6,1.351744964599609311e+01)(7,1.269108709335327134e+01)(8,1.219753433227539041e+01)(9,1.203074613571167006e+01)(10,1.189828987121581960e+01)(11,1.065349565505981388e+01)(12,1.074910181045532198e+01)(13,1.008184446334838924e+01)(14,1.007617784500122049e+01)(15,9.446936626434325746e+00)(16,9.645185031890868288e+00)(17,9.910988960266113423e+00)(18,8.991126537322998047e+00)(19,9.900570869445800781e+00)(20,8.735678224563597993e+00)(21,8.793057651519776030e+00)(22,8.765514087677001243e+00)(23,8.673707790374756144e+00)(24,8.574758272171020934e+00)(25,8.390970783233642649e+00)(26,8.788851041793822816e+00)(27,8.702681913375855061e+00)(28,8.640348482131958718e+00)(29,8.429963312149048704e+00)(30,8.524811229705811400e+00)(31,8.143984584808348970e+00)(32,8.186071977615355877e+00)(33,8.798317193984985352e+00)(34,8.112805919647216868e+00)(35,9.349773807525634695e+00)(36,8.362607660293578604e+00)(37,8.274409189224243732e+00)(38,8.312953987121581179e+00)(39,8.062538404464721253e+00)(40,8.162601242065429474e+00)(41,8.055403146743774556e+00)(42,8.050005245208740945e+00)(43,8.021660079956054901e+00)(44,7.820090818405151190e+00)(45,7.902752828598022639e+00)(46,7.880344486236571910e+00)(47,8.512783279418945526e+00)(48,7.748364706039428285e+00)(49,7.739994297027587677e+00)(50,7.760106449127197159e+00)};
\addplot[purple,line width=0.01mm, opacity=0.4]coordinates{(1,2.551590976333618244e+02)(2,2.719973917007446218e+01)(3,1.856670972824096566e+01)(4,1.669788095474243050e+01)(5,1.558893283843994126e+01)(6,1.450035492897033684e+01)(7,1.350346272468566866e+01)(8,1.270142608642578175e+01)(9,1.199951788902282779e+01)(10,1.132400588989257884e+01)(11,1.082919542312622063e+01)(12,1.038399621009826745e+01)(13,1.001973296165466287e+01)(14,9.722118349075318022e+00)(15,9.461969604492187713e+00)(16,9.253010740280151580e+00)(17,9.145256328582764382e+00)(18,8.878823375701903942e+00)(19,8.991133041381836222e+00)(20,8.398117542266845703e+00)(21,8.312121281623840119e+00)(22,8.190419192314147168e+00)(23,8.262139935493468812e+00)(24,8.128490624427795908e+00)(25,8.037466578483581259e+00)(26,8.019894466400145916e+00)(27,7.927411298751831481e+00)(28,7.894237818717956934e+00)(29,7.872856044769287465e+00)(30,7.804214062690735254e+00)(31,7.716501202583312846e+00)(32,7.737706055641174530e+00)(33,7.731075811386108221e+00)(34,7.629594621658324805e+00)(35,7.610565180778503525e+00)(36,7.631288108825683203e+00)(37,7.548831515312194540e+00)(38,7.515339131355285751e+00)(39,7.302114915847778498e+00)(40,7.337036209106445384e+00)(41,7.222091798782348526e+00)(42,7.200973534584045765e+00)(43,7.224963150024414027e+00)(44,7.211328105926513743e+00)(45,7.092880482673645126e+00)(46,7.154774851799011515e+00)(47,7.087131323814392481e+00)(48,7.006362247467040838e+00)(49,6.967289905548096129e+00)(50,6.921666340827941966e+00)};
\addplot[violet,line width=0.01mm, opacity=0.4]coordinates{(1,4.293977745056152173e+01)(2,2.050365894317626925e+01)(3,2.347365692138671989e+01)(4,1.709919361114501868e+01)(5,1.611479694366455107e+01)(6,1.433188743591308523e+01)(7,1.370870788574218757e+01)(8,1.353802297592163129e+01)(9,1.170534208297729428e+01)(10,1.162915355682373075e+01)(11,1.100233791351318402e+01)(12,1.033311212539672930e+01)(13,1.026439910888671925e+01)(14,1.019296909332275369e+01)(15,9.939277305603027912e+00)(16,9.477129611968994283e+00)(17,1.015518770217895472e+01)(18,9.803457889556884908e+00)(19,9.655607643127440909e+00)(20,9.093876199722290110e+00)(21,8.979881753921509002e+00)(22,8.753651304244995046e+00)(23,8.906185455322265909e+00)(24,8.739491252899169282e+00)(25,8.927640790939330273e+00)(26,9.515691928863525106e+00)(27,8.606152057647705078e+00)(28,8.728451051712035280e+00)(29,8.576707677841186594e+00)(30,8.737213430404663583e+00)(31,8.972852764129639169e+00)(32,8.531440868377686115e+00)(33,8.558304185867308789e+00)(34,8.532261228561401012e+00)(35,8.247324914932251616e+00)(36,8.407062702178954794e+00)(37,8.319399528503417685e+00)(38,8.637549686431885476e+00)(39,8.205482530593872781e+00)(40,8.108663024902343253e+00)(41,7.950946807861328125e+00)(42,7.995140533447266051e+00)(43,8.474848966598511169e+00)(44,7.941895837783813583e+00)(45,7.990514106750488565e+00)(46,8.029462547302246733e+00)(47,8.064659223556517986e+00)(48,8.139221067428588086e+00)(49,8.095211887359619496e+00)(50,7.821946916580199805e+00)};
\addplot[green,line width=0.01mm, opacity=0.4]coordinates{(1,2.786784500122070085e+02)(2,2.846051416397094869e+01)(3,1.837602048873901239e+01)(4,1.630463581085205149e+01)(5,1.511313911437988367e+01)(6,1.410995454788207937e+01)(7,1.310577877998352037e+01)(8,1.243253614425659137e+01)(9,1.163353864669799798e+01)(10,1.125014435768127363e+01)(11,1.068838168144226053e+01)(12,1.020190909385681088e+01)(13,9.957208328247070028e+00)(14,9.568582677841186879e+00)(15,9.441477203369140980e+00)(16,9.244711961746215678e+00)(17,8.811704797744750906e+00)(18,8.747290954589843182e+00)(19,8.532260646820068573e+00)(20,8.430342140197753409e+00)(21,8.345170459747315306e+00)(22,8.309248137474060769e+00)(23,8.280881052017212696e+00)(24,8.147205934524535564e+00)(25,8.072112402915955442e+00)(26,8.034486250877380442e+00)(27,7.982376775741577113e+00)(28,7.869739513397217223e+00)(29,7.832835502624512003e+00)(30,7.893211355209350799e+00)(31,7.914066977500915812e+00)(32,7.679824848175048402e+00)(33,7.655402908325195099e+00)(34,7.608539023399353063e+00)(35,7.674914417266845312e+00)(36,7.423851075172424707e+00)(37,7.330648698806762731e+00)(38,7.362579107284545898e+00)(39,7.205767154693603516e+00)(40,7.169507308006286550e+00)(41,7.171865839958190492e+00)(42,7.165641694068908940e+00)(43,7.050947833061218084e+00)(44,7.074324002265930389e+00)(45,7.025767602920532262e+00)(46,7.051692070960998926e+00)(47,7.083212528228759908e+00)(48,6.992894740104675044e+00)(49,6.983685021400451554e+00)(50,6.968180799484253285e+00)};
\addplot[OliveGreen,line width=0.01mm, opacity=0.4]coordinates{(1,4.774955612182617415e+01)(2,6.135180953979492102e+01)(3,1.861175071716308693e+01)(4,1.581850349426269453e+01)(5,1.486143840789794979e+01)(6,1.381893390655517528e+01)(7,1.273728675842285085e+01)(8,1.230268060684204023e+01)(9,1.143205017089843700e+01)(10,1.116231876373291065e+01)(11,1.074191175460815373e+01)(12,1.028364919662475607e+01)(13,9.973161945343017365e+00)(14,1.010956117630004947e+01)(15,1.001830663681030309e+01)(16,1.015091611862182575e+01)(17,9.279431209564208416e+00)(18,9.373521108627318910e+00)(19,9.292905693054200000e+00)(20,8.966108951568603658e+00)(21,8.866085519790649627e+00)(22,8.867432718276978321e+00)(23,8.659296150207518750e+00)(24,9.047463855743409056e+00)(25,8.619026126861571768e+00)(26,8.594057111740111665e+00)(27,8.750688419342040447e+00)(28,8.696527357101439648e+00)(29,8.566292610168456889e+00)(30,8.651014614105225320e+00)(31,8.609735736846923615e+00)(32,8.192158622741699148e+00)(33,8.424402637481689382e+00)(34,8.395176219940186257e+00)(35,8.428073301315308186e+00)(36,8.251064348220825906e+00)(37,8.621053953170775941e+00)(38,8.381355829238891886e+00)(39,8.035741291046143431e+00)(40,8.061946754455567188e+00)(41,8.177757406234741566e+00)(42,8.233399963378905895e+00)(43,7.993569002151489578e+00)(44,7.938720703125000000e+00)(45,7.899906482696533061e+00)(46,8.023820571899413778e+00)(47,8.077593994140624289e+00)(48,7.864949216842651580e+00)(49,8.055466861724854155e+00)(50,7.798602867126464666e+00)};
\addplot[purple,line width=0.01mm, opacity=0.4]coordinates{(1,2.705616918182373070e+02)(2,2.792845170974731417e+01)(3,1.839926038742065373e+01)(4,1.689777494430542149e+01)(5,1.565503451347351138e+01)(6,1.454752358436584458e+01)(7,1.361291218757629373e+01)(8,1.280917352676391552e+01)(9,1.199983971595764132e+01)(10,1.133865883827209409e+01)(11,1.087611063003540046e+01)(12,1.046684148788452084e+01)(13,9.990681333541870046e+00)(14,9.710790548324585103e+00)(15,9.498103733062743359e+00)(16,9.278225078582764240e+00)(17,9.041882843971253081e+00)(18,8.794804291725158762e+00)(19,8.783465871810912873e+00)(20,8.644667091369628409e+00)(21,8.565542917251587340e+00)(22,8.497506365776061443e+00)(23,8.369213781356812376e+00)(24,8.309839096069335795e+00)(25,8.209046092033386088e+00)(26,8.120170617103577015e+00)(27,7.841449217796325755e+00)(28,7.742594494819640794e+00)(29,7.725656104087829767e+00)(30,7.611893367767334162e+00)(31,7.641373248100280513e+00)(32,7.574974703788757147e+00)(33,7.580605015754700027e+00)(34,7.429040513038635218e+00)(35,7.403480739593505611e+00)(36,7.363154659271240021e+00)(37,7.284165039062499680e+00)(38,7.292521948814392196e+00)(39,7.266646771430969487e+00)(40,7.238029708862304901e+00)(41,7.201355996131897363e+00)(42,7.183455653190613077e+00)(43,7.101218638420104590e+00)(44,7.185594286918639995e+00)(45,7.053887300491332724e+00)(46,6.970415072441101145e+00)(47,6.965416007041930868e+00)(48,6.962700066566466894e+00)(49,6.917295112609862962e+00)(50,6.925958828926086497e+00)};
\addplot[violet,line width=0.01mm, opacity=0.4]coordinates{(1,4.707039459228515454e+01)(2,1.990381919860839943e+01)(3,1.801187744140624858e+01)(4,1.619487773895263771e+01)(5,1.640914623260498217e+01)(6,1.421906270980834996e+01)(7,1.345210077285766559e+01)(8,1.283234590530395458e+01)(9,1.210515422821044851e+01)(10,1.157089706420898523e+01)(11,1.064515357971191456e+01)(12,1.047326780319213846e+01)(13,1.047025047302246037e+01)(14,1.022309860229492173e+01)(15,9.847186813354491974e+00)(16,9.986494369506836222e+00)(17,9.522299900054932209e+00)(18,9.416505718231201527e+00)(19,9.439268121719360138e+00)(20,9.154892034530639933e+00)(21,9.385995502471923047e+00)(22,1.022454303741455028e+01)(23,8.694594421386717897e+00)(24,9.599595451354980824e+00)(25,9.020593910217284517e+00)(26,8.719929924011230682e+00)(27,8.515748357772826793e+00)(28,8.437906112670898295e+00)(29,8.489717273712157564e+00)(30,8.350541400909424539e+00)(31,8.583867454528808949e+00)(32,8.719503688812256570e+00)(33,8.560611858367920490e+00)(34,8.251308803558350391e+00)(35,8.335789747238159464e+00)(36,8.202313470840454812e+00)(37,8.069412136077881215e+00)(38,7.964565372467040660e+00)(39,8.048405427932738831e+00)(40,8.093617458343505433e+00)(41,7.941899757385254333e+00)(42,8.225135984420775870e+00)(43,9.272130146026611897e+00)(44,8.023134059906006144e+00)(45,8.072333278656005362e+00)(46,7.943795366287231374e+00)(47,7.915306053161621058e+00)(48,8.030414123535155824e+00)(49,7.902394561767578374e+00)(50,7.896928510665893519e+00)};
\addplot[green,line width=0.01mm, opacity=0.4]coordinates{(1,2.545863011550903252e+02)(2,2.734038108825683722e+01)(3,1.832438150405883803e+01)(4,1.632152922630309888e+01)(5,1.514005230903625510e+01)(6,1.397410150527954187e+01)(7,1.297820368766784682e+01)(8,1.226394248962402322e+01)(9,1.163954753875732351e+01)(10,1.091956818580627520e+01)(11,1.070832553863525405e+01)(12,1.009734252929687415e+01)(13,9.894472379684447816e+00)(14,9.490499782562256570e+00)(15,9.373030261993408274e+00)(16,9.119382400512694886e+00)(17,8.861036357879639169e+00)(18,8.791506729125977415e+00)(19,8.636036825180053000e+00)(20,8.616369619369507404e+00)(21,8.448905997276305868e+00)(22,8.358654398918151429e+00)(23,8.312997307777404643e+00)(24,8.384498777389525870e+00)(25,8.164817733764648366e+00)(26,7.944275751113891282e+00)(27,7.734351906776428365e+00)(28,7.712562317848205673e+00)(29,7.731607131958007528e+00)(30,7.669748048782349059e+00)(31,7.687588834762573597e+00)(32,7.561822786331176438e+00)(33,7.575699505805968847e+00)(34,7.504524083137511958e+00)(35,7.444376120567321564e+00)(36,7.297270307540893342e+00)(37,7.254910454750061355e+00)(38,7.280593543052673589e+00)(39,7.176109046936034908e+00)(40,7.093232517242431534e+00)(41,7.085007476806640447e+00)(42,6.983671898841858017e+00)(43,6.995260362625121964e+00)(44,7.029430923461913672e+00)(45,7.044995646476746032e+00)(46,6.952632150650024379e+00)(47,6.920285935401916610e+00)(48,6.905921669006347408e+00)(49,6.891732015609741246e+00)(50,6.834059576988220641e+00)};
\addplot[OliveGreen,line width=0.01mm, opacity=0.4]coordinates{(1,4.272142547607421648e+01)(2,2.041548213958740376e+01)(3,2.332688186645507855e+01)(4,2.738097227096557518e+01)(5,1.473594701766967852e+01)(6,1.950872110366821133e+01)(7,1.254012407302856502e+01)(8,1.193161682128906165e+01)(9,1.136191778182983469e+01)(10,1.190960725784301744e+01)(11,1.041300235748290959e+01)(12,1.063902065277099673e+01)(13,1.041701292037963889e+01)(14,1.018340167999267543e+01)(15,9.557617301940917187e+00)(16,9.219955453872680451e+00)(17,9.465586643218994567e+00)(18,9.177710323333739595e+00)(19,9.573772735595703409e+00)(20,8.946040792465209890e+00)(21,9.643827838897705007e+00)(22,8.623588237762451314e+00)(23,9.387292785644531179e+00)(24,9.073842735290527983e+00)(25,1.049850969314575266e+01)(26,8.385433654785156676e+00)(27,8.344998807907105132e+00)(28,9.203847885131835938e+00)(29,8.685812225341797088e+00)(30,8.188591051101685281e+00)(31,8.095267076492309144e+00)(32,8.088376092910767312e+00)(33,8.106200799942016033e+00)(34,8.499477396011352326e+00)(35,8.221996364593506357e+00)(36,7.856120653152466105e+00)(37,8.059570760726929350e+00)(38,8.077922458648680859e+00)(39,7.997269744873046982e+00)(40,7.950729703903197887e+00)(41,7.841861495971679474e+00)(42,7.816258659362793182e+00)(43,7.726844778060913299e+00)(44,7.814646034240722905e+00)(45,7.916554269790649023e+00)(46,7.788899374008178889e+00)(47,7.732759208679198970e+00)(48,7.837776823043823171e+00)(49,7.840499629974365448e+00)(50,7.707770290374756250e+00)};
\addplot[purple]coordinates{(1,257.3879475)(2,27.5101331)(3,18.46158466)(4,16.60416695)(5,15.41853145)(6,14.3692435)(7,13.50498283)(8,12.7232302)(9,11.9968618)(10,11.3905121)(11,10.92898211)(12,10.47190513)(13,10.08708388)(14,9.780405478)(15,9.572810404)(16,9.296413406)(17,9.127084713)(18,8.910969335)(19,8.816743819)(20,8.587293357)(21,8.454476196)(22,8.382255224)(23,8.305871185)(24,8.268458206)(25,8.192098483)(26,8.112884014)(27,7.998938662)(28,7.894200067)(29,7.855071735)(30,7.760234271)(31,7.750485023)(32,7.655720898)(33,7.664893161)(34,7.620772579)(35,7.571351175)(36,7.454688787)(37,7.389286237)(38,7.388808868)(39,7.320116116)(40,7.294823841)(41,7.21940625)(42,7.169773458)(43,7.148763561)(44,7.143613891)(45,7.077957354)(46,7.05325021)(47,7.041337225)(48,7.01039761)(49,6.95436717)(50,6.923133492)};
\addplot[violet]coordinates{(1,45.44369049)(2,20.45850488)(3,22.01633967)(4,24.99879668)(5,16.50265433)(6,14.33202073)(7,13.6049197)(8,12.78793055)(9,12.03347376)(10,11.58466008)(11,10.98748235)(12,10.4734044)(13,10.39175127)(14,10.1280362)(15,9.777622608)(16,9.713095413)(17,9.809038509)(18,9.670449383)(19,9.4018302)(20,9.183821365)(21,9.145810791)(22,9.235035164)(23,9.028806517)(24,8.899166889)(25,8.793612818)(26,8.914081263)(27,8.693158579)(28,8.508458416)(29,8.578722967)(30,8.50706138)(31,8.624466143)(32,8.451798201)(33,8.495652235)(34,8.484553516)(35,8.478515314)(36,8.215509548)(37,8.403542788)(38,8.250296389)(39,8.166927858)(40,8.216566965)(41,8.121053604)(42,8.182255775)(43,8.362776323)(44,7.998373859)(45,8.029089037)(46,7.932551937)(47,7.959324457)(48,7.979170996)(49,8.004445923)(50,7.86243568)};
\addplot[green]coordinates{(1,269.1180581)(2,28.21265573)(3,18.40226598)(4,16.42302675)(5,15.21275809)(6,14.13723531)(7,13.14906785)(8,12.39609316)(9,11.69206872)(10,11.11319239)(11,10.64653439)(12,10.21312326)(13,9.90192655)(14,9.567035707)(15,9.401223558)(16,9.163649295)(17,8.925355853)(18,8.807712996)(19,8.657624128)(20,8.591970632)(21,8.452182244)(22,8.28339433)(23,8.239628017)(24,8.149788473)(25,8.107046533)(26,7.95557965)(27,7.836849302)(28,7.81011553)(29,7.697173212)(30,7.641633639)(31,7.646899995)(32,7.605880239)(33,7.543636947)(34,7.46002098)(35,7.466055063)(36,7.349977418)(37,7.27898359)(38,7.269490557)(39,7.147420347)(40,7.09822743)(41,7.108537794)(42,7.081504741)(43,7.037955232)(44,7.043813616)(45,7.017336998)(46,7.006936838)(47,6.996316055)(48,6.937507799)(49,6.924841831)(50,6.897511486)};
\addplot[OliveGreen]coordinates{(1,47.50182178)(2,30.68728327)(3,19.68424953)(4,18.29376094)(5,15.03333042)(6,15.33791863)(7,12.89367659)(8,12.2843216)(9,11.94397123)(10,11.56019573)(11,10.73157223)(12,10.5044645)(13,10.12754359)(14,10.38881053)(15,9.809634464)(16,9.612795427)(17,9.718599434)(18,9.197471745)(19,9.753666145)(20,9.072291718)(21,9.236122705)(22,8.766421022)(23,8.788898129)(24,8.958717487)(25,9.128627947)(26,8.551600418)(27,8.563221022)(28,8.688926815)(29,8.473807009)(30,8.366534389)(31,8.358534201)(32,8.182396103)(33,8.396462811)(34,8.271012892)(35,8.415828848)(36,8.167897335)(37,8.25534914)(38,8.181384558)(39,8.027670786)(40,8.060003098)(41,7.99971981)(42,7.990061356)(43,7.941530552)(44,7.85889572)(45,7.919655016)(46,7.905162687)(47,8.018190882)(48,7.85454203)(49,7.855816732)(50,7.799603132)};
\end{axis}
\end{tikzpicture}
\caption{The graph shows a difference between the usage of the standard hypercolumn technique and the hypercolumn with stairstep in the term of the loss. The thin lines denote particular learning runs, and the thick lines are mean of the runs.}
\label{fig-graph-hc-stairstep}
\end{figure}
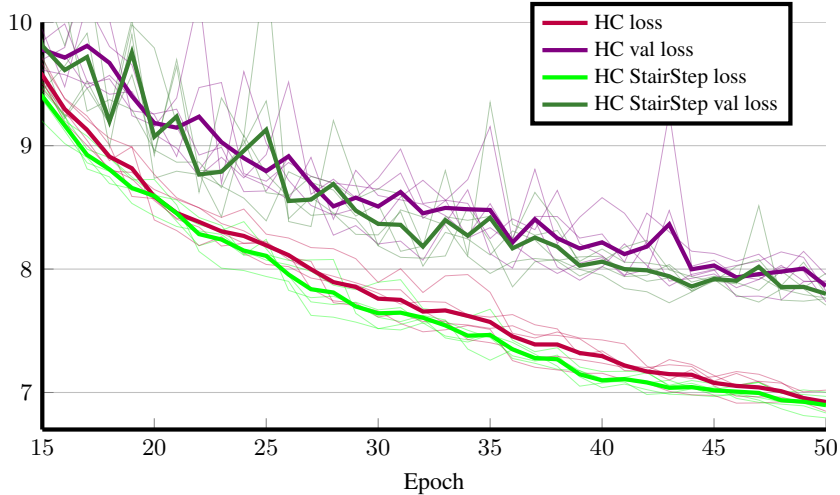

For understanding the practical impact, we initiated the following experiment. We trained Poly-YOLO for 200 training and 100 validation images from Cityscapes dataset~\cite{cordts2016cityscapes} for the version with direct upscaling and stairstep upscaling used in the hypercolumn. We ran the training process five times for each of the versions and plotted the training progress in the graph in Figure~\ref{fig-graph-hc-stairstep}. The graph shows that the difference is tiny, but it is evident that stairstep interpolation in hypercolumn yields slightly lower training and validation loss. The improvement is obtained for the identical computation time.

\noindent

%\begin{table}[!ht]
%% increase table row spacing, adjust to taste
%\renewcommand{\arraystretch}{1.3}
%\caption{Comparison of YOLO versions}
%\label{table^example}
%\centering
%\begin{tabular}{|p{32mm}|ccc|}
%\hline
%&YOLOv2&YOLOv3&Poly-YOLO\\
%\hline
%Backbone&Darknet-19&Darknet-53&SE-Darknet-53\\
%Output scales&1&3&1\\
%Anchors per scale&5&3&9\\
%Grid dividers&?&8,16,32&4\\
%Anchor distribution problem&no&yes&no\\
%Parameters&?&?&37M\\
%\hline
%\end{tabular}
%\end{table}

%\begin{figure}[!ht]
%    \includegraphics[width=88mm]{imgs/se^extension.pdf}
%\caption{\color{red}todo honza: prekreslit do vektorove %grafiky.\color{black}Left: original YOLO residuated block. Right: The block %enhanced by squeeze-excitation technique with \emph{standard} placing of the SE %block.}
%\label{se-extension}
%\end{figure}

The last way how we propose to modify YOLO's architecture is the usage of squeeze-and-excitation (SE) blocks~\cite{hu2018squeeze} in the backbone. Darknet-53, like many other neural networks, uses repetitive blocks, where each block consist of coupled convolutions with residual connection. The squeeze-and-excitation blocks allows the usage of both spatial and channel-wise information, which leads to accuracy improvement. %In~\cite{hu2018squeeze}, authors propose several ways, how these blocks can be integrated into existing architectures. They find that the best performance is achieved for the version SE-PRE. Because coupled convolutions of Darknet-53 are special (it consist of bottleneck convolution followed by an expansion convolution), we realized ablation study for SE-Standard, SE-PRE, SE-POST, and SE-Identity. We have achieved the best results for the version SE-Standard while SE-PRE was runner-up. For the detailed results, see Figure~\ref{fig-graph-se-type} in Discussion section. 
By the addition of squeeze-and-excitation blocks and by working with higher output resolution, computation speed is decreased. Because speed is the main advantage of YOLO, we reduced the number of convolutional filters in the feature extraction phase. Namely, it is set to 75\% of the original number. Also, the neck and head are lighter, together having 37.1M parameters, which is significantly less than has YOLOv3 (61.5M). Still, Poly-YOLO achieves higher precision than YOLOv3 -- see Section~\ref{subsec-results}. We also propose Poly-YOLO lite, which is aimed at higher processing speed. In the feature extractor and the head, this version has only 66\% of filters of Poly-YOLO. Finally, $s_1$ is reduced to 1/4. The number of parameters of Poly-YOLO lite is 16.5M.

We want to highlight that for feature extraction, an arbitrary SOTA backbone such as (SE)ResNeXt~\cite{xie2017aggregated} or EfficientNet~\cite{tan2019efficientdet} can be used, which would probably increase the overall accuracy. Such an approach can also be seen in the paper YOLOv4~\cite{bochkovskiy2020yolov4}, where the authors use a different backbone and several other tricks (that can also be applied in our approach) but the head of the original YOLOv3 is left unchanged. The issues we described and removed in Poly-YOLO actually arise from the design of the head of YOLOv3, and a simple swap of a backbone will not solve them. The model would still suffer label rewriting and improper anchor distribution. In our work, we have focused on performance improvement achieved by conceptual changes and not brute force. Such improvements are then widely applicable, and a modern backbone can be easily integrated.

\section{Instance segmentation with Poly-YOLO}
\begin{figure*}[!t]
    \centering
    \includegraphics[width=26mm]{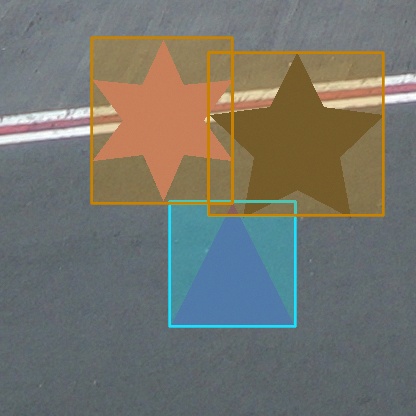}
    \includegraphics[width=26mm]{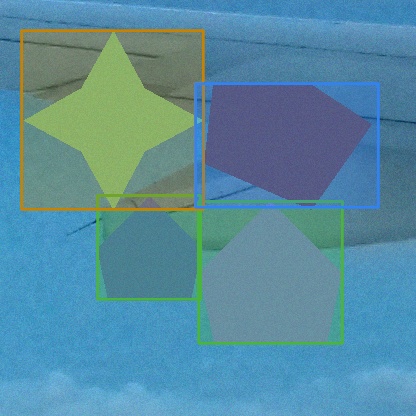}
    \includegraphics[width=26mm]{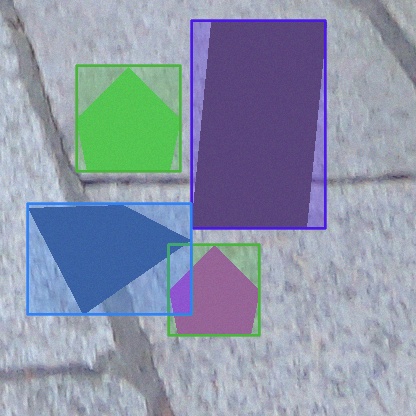}
    \includegraphics[width=26mm]{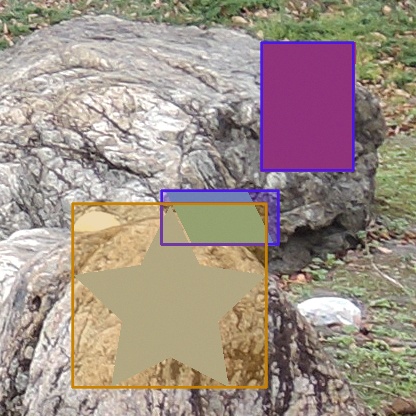}
    \includegraphics[width=26mm]{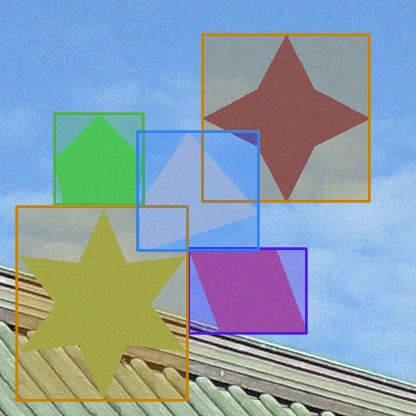}
    \includegraphics[width=26mm]{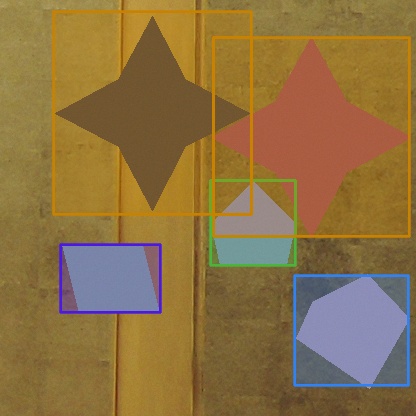}\\[1mm]
    \includegraphics[width=26mm]{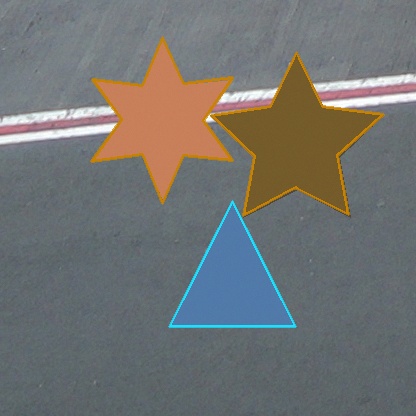}
    \includegraphics[width=26mm]{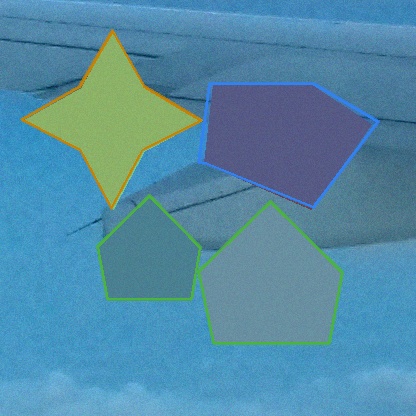}
    \includegraphics[width=26mm]{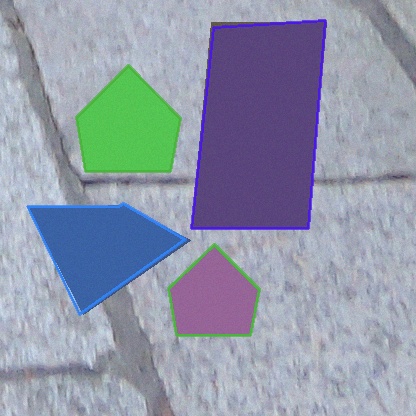}
    \includegraphics[width=26mm]{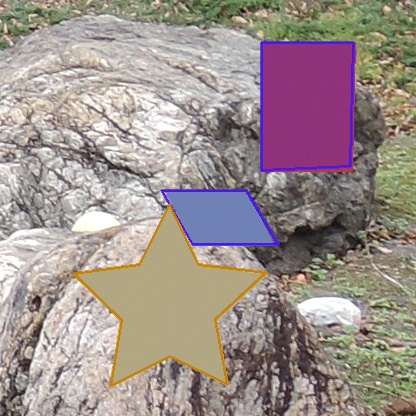}
    \includegraphics[width=26mm]{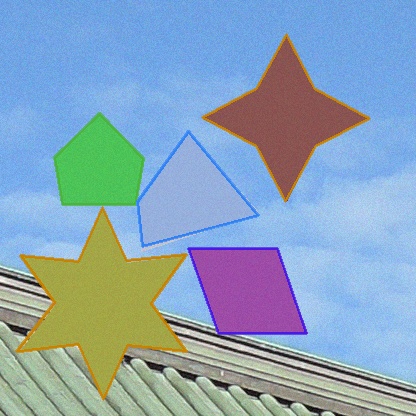}
    \includegraphics[width=26mm]{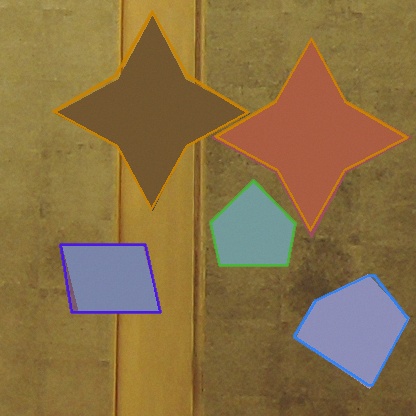}\\
\caption{Comparison of Poly-YOLO bounding box detection (top) and Poly-YOLO bounding polygon detection (bottom).}
\label{fig-precision-of-bounds}
\end{figure*}

The last sentence in YOLOv3 paper~\cite{redmon2018yolov3} says "\emph{Boxes are stupid anyway though, I’m probably a true believer in masks except I can’t get YOLO to learn them.}" Here, we show how to extend YOLO with masking functionality (instance segmentation) without a big negative impact on its speed. In our previous work~\cite{hurtik2020yoloasc}, we were focusing on more precise detection of YOLO by means of irregular quadrangular detection. We proved that the extension for quadrangular detection converges faster. We also demonstrated that classification from quandrangular approximation yields higher accuracy than from rectangular approximation. The limitation of that approach lies in the fixed number of detected vertices, namely four. Here, we introduce a polygon representation that is able to detect objects with a varying number of vertices without the usage of a recurrent neural network that would slow down the processing speed. To see a practical difference between the quality of bounding-box detection and polygon-based detection, see Figure~\ref{fig-precision-of-bounds}, where we show results from Poly-YOLO trained to detect various geometric primitives including random polygons.

\begin{figure}[!h]
    \centering
    \includegraphics[width=110mm]{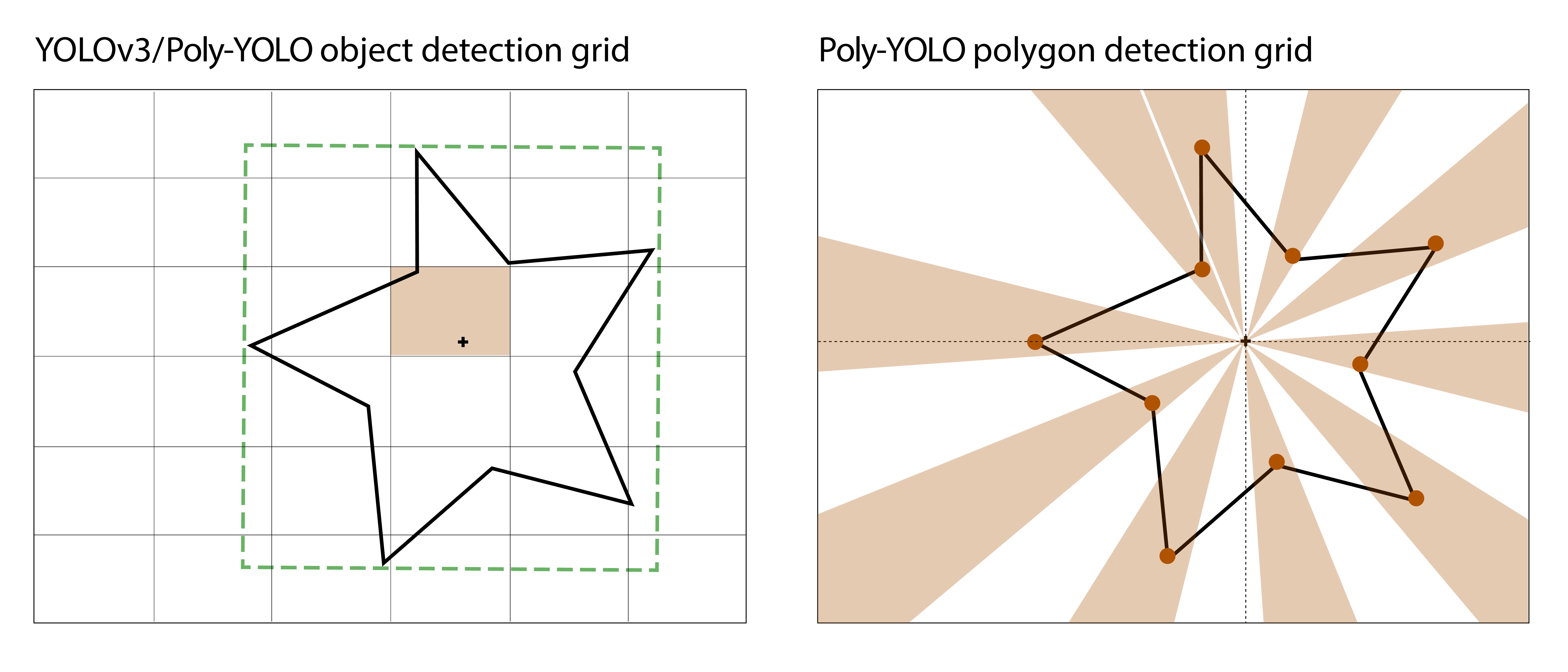}
\caption{The image illustrates grids used in Poly-YOLO. Left: the rectangular grid, which is taken from YOLOv3. A cell where an object's bounding box has its center predicts its bounding box coordinates. Right: the grid based on circular sectors used in Poly-YOLO for the detection of vertices of the polygon. The center of the grid coincides with the center of the object's bounding box. Each circular sector is then responsible for detecting polar coordinates of the particular vertex. Sectors, where no vertex is present, should yield confidence equal to zero.}
\label{fig-polygon^detection}
\end{figure}

\subsection{The principle of bounding polygons}
YOLOv3 uses a perpendicular grid consisting of cells where each cell can detect a bounding box, or bounding boxes in the case of multiple anchors. We extend each cell with an additional polar sub-grid, see  Figure~\ref{fig-polygon^detection}. Let us recall, we can describe a box as $\mathbf{b}_i=(b_i^{x^1}, b_i^{y^1}, b_i^{x^2}, b_i^{y^2})$, i.e., as a tuple of its top-left and bottom-right coordinates. We propose to extend the tuple as $\mathbf{b}_i=(b_i^{x^1}, b_i^{y^1}, b_i^{x^2}, b_i^{y^2}, V_i)$, where $V_i=\{\mathbf{v}_{i,0}, \dots, \mathbf{v}_{i,n}\}$ is a set of polygon vertices of a given object for $n$ polar cells. Furthermore, $\mathbf{v}_{i,j}=(\alpha_{i,j}, \beta_{i,j}, \gamma_{i,j})$, where $\alpha$ and $\beta$ are coordinates of a polygon vertex in a polar coordinate system and $\gamma$ is its confidence. If no vertex is present in a polar cell, the confidence should be ideally equal to zero, otherwise it should be equal to one.

In a common dataset, many objects are masked with a similar shape because they are captured from a similar viewpoint; the difference is only in the object size. For example, instances of car plates, hand gestures, humans, or cars have almost identical shapes. The general shape can be easily described using polar coordinates, which is the motivation why we use the polar coordinate system instead of the cartesian one for bounding polygons. Here, $\alpha_{i,j}$ stands for the distance of a vertex from the origin and $\beta_{i,j}$ for an oriented angle. The center of a bounding box is used as the origin. %For the center lies in a cell that detects it. Furthermore, it always holds that all vertices of the polygon lie inside of the bounding box. Using this fact, 
Furthermore, we divide the vertex distance from the origin by the length of a diagonal of the bounding box in order to obtain $\alpha_{i,j}\in[0,1]$. Then, during inference, when the bounding box with a bounding polygon is detected, the absolute distance from the origin is obtained by multiplying $\alpha_{i,j}$ with a diagonal of the detected box. The principle allows the network to learn general, size-independent shapes, not particular instances with sizes. For example, let us suppose two images of the same car placed in two distinct distances from a camera so that in each image, its size will be different. The model will be trained to detect confidence, angles, and relative distances from the bounding box center for every vertex. These values will be the same for both images. When predictions are realized, the distances are multiplied by the box diagonal, and the particular values of the two differently sized cars will be obtained. In comparison with PolarMask~\cite{xie2019polarmask} which has to predict distinct distances for different object sizes,  this sharing of values should make the learning easier.

Still, additional improvement is possible. For the oriented angle it holds that $\beta_{i,j}\in[0,360]$, which can be changed to $\beta_{i,j}\in[0,1]$ by a linear transformation. Because our polar system is split into polar cells, it would be beneficial to focus the inside of a cell to a particular part of the angle interval, which is covered by the cell. When a certain polar cell fires with high confidence, the vertex has to be inside the cell. Therefore, we propose to take $\beta_{i,j}\in[\beta^1,\beta^2]$, where $\beta^1_{i,j}$ and $\beta^2_{i,j}$ are the minimum, and the maximum angle captured by a polar cell in which the vertex lies. Then, we make linear transformation of $\beta_{i,j}$, where $\beta^1_{i,j}= 0$ and $\beta^2_{i,j}=1$ holds. In other words, when a certain polar cell has high confidence, we know that it contains a vertex. By a distance from the origin and the location of the polar cell, we know its approximate position, and by the angle inside the cell, we refine the position precisely.% For practical demonstration, let us suppose we have 20 polar cells for bounding polygon. If a network has precision $\pm 0.1$ in the first case when the whole interval is taken, the angle will be searched within the precision of $\pm36_\circ$, while in the second case within the precision of $\pm 1.8_\circ$.

\subsection{Integration with Poly-YOLO}
\label{subsec-integration-with-yolo}
The idea of detecting bounding polygons is universal and can be easily integrated into an arbitrary neural network. In general, three parts have to be modified: the way how data are prepared, an architecture, and a loss function. For extraction of bounding polygons from semantic segmentation labels, see Section~\ref{ref-subsec-preparing-data}. The extracted bounding polygons have to be augmented in the same way as data for bounding boxes.

The architecture has to be modified to produce the intended values. In the case of Poly-YOLO, the number of convolutional filters in the output layer has to be updated. When we detect only bounding boxes, the last layer is represented by $n=n^a(n^c+5)$ convolutional filters with a kernel of dimension $(1, 1)$, where $n^a$ is a number of anchors (nine, in our case) and $n^c$ stands for a number of classes. After integrating the extension for polygon-based object detection, we obtain $n=n^a(n^c+5+3n^v)$, where $n^v$ is a maximal number of detected vertices per polygon. We can observe that $n^v$ has a high impact on the number of convolutional filters. For example, when we have nine anchors, twenty classes, and thirty vertices, the output layer detecting bounding-boxes and polygons will have $4.6\times$ more filters than when detecting bounding-boxes only. On the other hand, the increase happens only in the last layer; all remaining YOLO layers have the same number of parameters. From that point of view, the total number of the NN parameters is increased by a negligible 0.83\%, and the processing speed is not affected. The weak point lies in the fact that the increase is in the last layer, which processes high-resolution feature maps. This causes an increased demand for VRAM for a symbolic tensor when the network is trained, which may lead to a decrease of the maximum possible batch size used during the learning phase.

For explaining how a loss function has to be modified, we describe the multi-part loss function $\ell$ used in Poly-YOLO as follows:
\begin{align*}
\ell=\sum^{G^wG^h}_{i=0}\sum^{n^a}_{j=0}& q_{i,j}\Big[\ell_1(i,j) + \ell_2(i,j) + \ell_3(i,j) + \ell_5(i,j)\Big]+\ell_4(i,j),
\end{align*}
where $\ell_1(i,j)$ is a loss for a prediction of a center of a bounding box, $\ell_2(i,j)$ is a loss for dimensions of a box, $\ell_3(i,j)$ is confidence loss, $\ell_4(i,j)$ is class prediction loss and $\ell_5(i,j)$ is a loss for a bounding polygon made of distance, angle, and vertex confidence prediction. Finally, $q_{i,j} \in \{0,1\}$ is a constant indicating if the $i$-th cell and the $j$-th anchor contains a label or not. The loss iterates over $G^wG^h$ grid cells and $n^a$ anchors. The parts $\ell_1, \dots, \ell_4$ are taken from YOLOv3 and modified into a form working with a single output layer.  The part $\ell_5$ is new and extends Poly-YOLO with the functionality of polygon detection. In the following formulas, we use $\widehat{\cdot}$ to denote predictions of the network. The parts of the loss function are defined as follows:

\begin{equation*}
 \ell_1(i,j) = z_{i,j} \left[H(c^x_{i,j},\widehat{c}^x_{i,j})+H(c^y_{i,j},\widehat{c}^y_{i,j})\right],   
\end{equation*}
where $c^x_{i,j}$ and  $c^y_{i,j}$ are coordinates of the center of a box, $H(\cdot, \cdot)$ is binary crossentropy, $z_{i,j}= 2 - w_{i,j}h_{i,j}$ serves for a relative weighting of $(i,j)$-th box size according to its width $w_{i,j}$ and height $h_{i,j}$.
\begin{equation*}
 \ell_2(i,j) = 0.5 z_{i,j} \Bigg[\Bigg.\left(log\left(\frac{w_{i,j}}{a_j^w}\right) - \hat{w}_{i,j}\right)^2 + 
 \left(log\left(\frac{h_{i,j}}{a_j^h}\right) - \hat{h}_{i,j}\right)^2\Bigg.\Bigg],  
\end{equation*}
where $a_j^w$ and $a_j^h$ are width and height of the $j$-th anchor.
\begin{equation*}
 \ell_3(i,j) = q_{i,j}H(q_{i,j},\hat{q}_{i,j}) + (1-q_{i,j})H(q_{i,j},\hat{q}_{i,j})I_{i,j},   
\end{equation*}
where $\hat{q}_{i,j}$ is predicted confidence and $I_{i,j}$ is a mask which excludes the part of a loss for the $i$-th cell if $q_{i,j} = 0$ but its prediction has IoU$>$0.5.
\begin{equation*}
 \ell_4(i,j) = \sum^{c}_{k=0}H(C_{i,j,k}-\hat{C}_{i,j,k}),   
\end{equation*}
where $C^{i,j,k}$ is $k$-th class probability in $i$-th cell. Finally,
\begin{equation*}
\label{eq:polygon^loss}
 \ell_5(i,j) = 0.2\sum^{v}_{l=0} z_{i,j}\Bigg[\Bigg.\gamma_{i,j,k} \left(log\left(\frac{\alpha_{i,j,k}}{a_j^d}\right)-\hat{\alpha}_{i,j,k}\right)^2
+\gamma_{i,j,k}H(\beta_{i,j,k}, \hat{\beta}_{i,j,k}) + H(\gamma_{i,j,k}, \hat{\gamma}_{i,j,k})\Bigg.\Bigg],   
\end{equation*}
where $a_j^d$ is the diagonal of the $j$-th anchor. Note that the last equation is our polygon representation loss, one of our main contributions.

The described scheme of integration results in the simultaneous detection of both bounding boxes and bounding polygons. Such combination may be beneficial due to synergy -- convolutional neural networks detect edges in its bottom, then combine them into more complex shapes in the middle and propose highly-descriptive abstract features in head~\cite{lee2011unsupervised}. Because the polygon vertices always lie in the bounding box and because the vertices delimit the same object as the bounding box, the intuition is the bounding polygon part will find features useful for bounding box and vice versa. The assumption is that the training of YOLO with polygon shape detection extension will be more efficient and converge faster. The principle is well known and described in the literature as \emph{Auxiliary task learning}~\cite{ruder2017overview}. For completeness, let us suppose a special case when an object is a perpendicular box. For such a case, the contour of the bounding box will coincide with the contour of the bounding polygon, and the left-top vertex will be detected by both the bounding box and polygon. Still, the two detections will be synergistic, and the training will require a shorter time than the training of vanilla bounding box detection. For the verification of the claims, see results of Poly-YOLO detection with/without bounding polygon detection in Section~\ref{subsec-results}.

\section{Benchmarks}
\label{sec-benchmark}
Here, we describe the experiments and results which we realized. We divided them into two scenarios, algorithms for pure bounding box detection and algorithms for instance segmentation. Each scenario includes three datasets, namely Simulator, Cityscapes, and IDD. For the training, we use three computers with RTX2080Ti, RTX2060, or GT1080 graphics card, i.e., the mid-tier graphics cards. The inference time is always measured on the computer with the RTX2080Ti.

\subsection{Preparing data for Poly-YOLO}
\label{ref-subsec-preparing-data}
As was stated before, one of the features of Poly-YOLO is an object detection with the ability to estimate a polygon tightly wrapping around the object. For that purpose, quite specific data must be fetched in. Commonly, publicly available data sets focus on pixel-precise segmentation masks. The mask assigns each pixel to one of the many predefined classes. Unfortunately, for Poly-YOLO, we need a polygonal representation and not pixel-wise representation. Because of that, some pre-processing became inevitable. 

As an input to the extraction of bounding polygon, we suppose a blob of pixel coordinates of a given object. That notation is standard for general pixel segmentation tasks. Then, we find the most distant points from an object center. It means that if an object is folded, we do not extract the inner boundary points, just the most distant ones. For example, an object with a shape of a swiss roll will have a contour similar to the circle. Finally, we erase points that lie in a straight line between two other points. Note, for a single object, we always extract a single bounding polygon. If we take, e.g., a car which is partly overlapped by a tree, the bounding polygon will bound the whole car, including the part covered behind the tree.

Another way to obtain training data is to generate them synthetically. For that, we use two of our tools. The first one serves for a generation of complex and realistic scenes, as is shown in Figure~\ref{fig-simulator}. The second one can generate an infinite amount of images, where the following parameters can be configured: the resolution of images, the number of geometric primitives per image, the type of geometric primitives, the range of their size. It is also possible to add a random background. For the illustration, see Figure~\ref{fig-precision-of-bounds}. The tool is available at our GitLab repository\footnote{gitlab.com/irafm-ai/poly-yolo/-/tree/master/synthetic\_dataset}.

\subsection{Datasets}
In the benchmark, we use three datasets: Simulator, Cityscapes~\cite{cordts2016cityscapes}, and India Driving~\cite{varma2019idd}. 

\begin{figure}[!h]
    \centering
    \includegraphics[width=130mm]{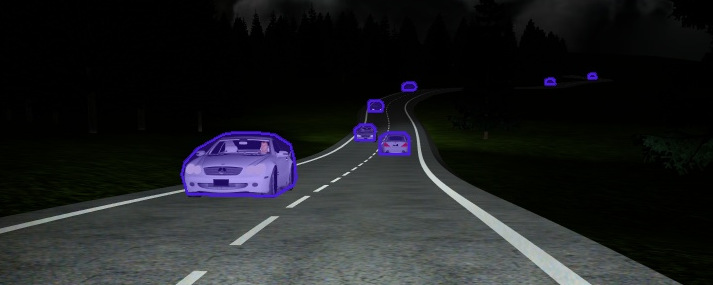}
    \caption{The figure shows an illustrative image from the Simulator dataset. The blue color shows prediction realized by Poly-YOLO lite at 52FPS. The image was slightly cropped to increase visibility.}
    \label{fig-simulator}
\end{figure}

Simulator is our own synthetic dataset available online\footnote{gitlab.com/irafm-ai/poly-yolo/-/tree/master/simulator\_dataset} consisting of 700 training, 90 validation, and 100 test images with a resolution of $600\times800$px. The dataset is useful for fast prototyping, hyperparameter searching, or as a starting point for transfer learning because the low number of images allows fast training, and the captured scenes are trivial. It includes only a single class (a car), where its particular instances are rendered using a single 3D model. On the other hand, the scene is illuminated by physically-precise lights. An illustrative image with detections by Poly-YOLO are visualized in Figure~\ref{fig-simulator}.

\noindent
\begin{figure*}[!h]
\small
\centering
\begin{tikzpicture}
      \begin{axis}[
        height=45mm,width=90mm,
        ybar,
        enlargelimits=0.07,
        xlabel={Maximal number of vertices per polygon},
        ymin=0,
        ymax=12,
        axis x line*=bottom,
        axis y line*=left,
        symbolic x coords={8, 12, 18, 24, 36, 60, 90},
        xtick=data,
        nodes near coords, 
    	nodes near coords align={vertical},
        x tick label style={rotate=0,anchor=center, font=\footnotesize},
        every node near coord/.append style={font=\footnotesize},
        legend style={at={(0.5,1.1)},anchor=north,  font=\footnotesize,nodes={scale=0.9, transform shape}},legend columns=2,
        ]
        \addplot[fill=blue!40!white]coordinates{(8,5.4)(12,5.8)(18,6.5)(24,6.9)(36,7.7)(60,8.8)(90,9.6)};\addlegendentry{Loss~~~~}
        \addplot[fill=orange!40!white]coordinates{(8,6.5)(12,6.9)(18,7.5)(24,7.8)(36,8.5)(60,9.5)(90,10.3)};\addlegendentry{Val loss}
      \end{axis}
    \end{tikzpicture}
~
\begin{tikzpicture}
      \begin{axis}[
        height=45mm,width=90mm,
        ybar,
        enlargelimits=0.07,
        xlabel={Number of anchors},
        ymin=0,
        ymax=12,
        axis x line*=bottom,
        axis y line*=left,
        symbolic x coords={1, 3, 6, 9, 12, 15, 18},
        xtick=data,
        nodes near coords, 
    	nodes near coords align={vertical},
        x tick label style={rotate=0,anchor=center, font=\footnotesize},
        every node near coord/.append style={font=\footnotesize},
        legend style={at={(0.5,1.1)},anchor=north,  font=\footnotesize,nodes={scale=0.9, transform shape}},legend columns=2,
        ]
        \addplot[fill=blue!40!white]coordinates{(1,7.1)(3,6.9)(6,6.8)(9,6.9)(12,7.0)(15,7.0)(18,7.2)};\addlegendentry{Loss~~~~}
        \addplot[fill=orange!40!white]coordinates{(1,8.2)(3,8.0)(6,7.7)(9,7.8)(12,8.0)(15,7.9)(18,8.0)};\addlegendentry{Val loss}
      \end{axis}
    \end{tikzpicture}
    \vspace{-3mm}
\caption{Left: Dependence of number of vertices on loss for 9 used anchors. Right: Dependence of number of anchors on loss for 24 used vertices.}
\label{graph-nr-vertices}
\end{figure*}
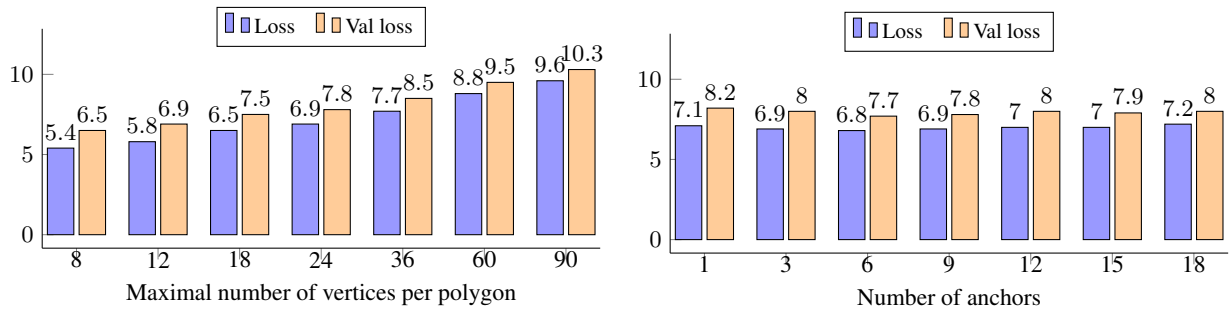

Cityscapes is a dataset captured from a car driven through various German cities. The images were captured during the day or evening, and all of them have the same resolution of 2048$\times$1024px. For the Cityscapes, both bounding boxes and pixel-level labels are available. The process, how we extracted polygons from pixel-level annotations is described in Section \ref{ref-subsec-preparing-data}. Notwithstanding, the pixel-level labels include objects such as sky, building, tree, etc., we use only objects connected with traffic such as a car, pedestrian, or bike. In total, we predict 12~classes. Because the testing dataset does not contain labels, we moved several images from the training to the testing dataset. Finally, our train/valid/test datasets consist of 2474/500/500 images.

IDD is a dataset similar to Cityscapes that focuses on unstructured traffic on India's roads. In contrast to Cityscapes, road borders captured in the images are fuzzy, traffic is heavier and messy, and images have various resolutions. The distribution of images is as follows -- 14010 for training, 977 for validation, and 1049 for testing.

\begin{table*}[!ht]
\scriptsize
\renewcommand{\arraystretch}{1.1}
\caption{The results of the involved algorithms for bounding box detection and instance segmentation on the three datasets.}
\label{tab-all-results}
\centering
\begin{tabular}{p{15mm}p{21mm}ll|c|ccc|ccc|c}
\hline
 & &&&\textbf{Instance}& & \textbf{Box} && &  \textbf{Mask} & \\
\textbf{Method} & \textbf{Backbone} &\textbf{Resolution}~& \textbf{~\#Parameters}&\textbf{segment.}& ~~\textbf{AP} & \textbf{AP$_{50}$}~&\textbf{AP$_{75}$}~~& ~~\textbf{AP}~&\textbf{AP$_{50}$}~&\textbf{AP$_{75}$}~~&~~\textbf{FPS}~~\\
\hline
\hline
\multicolumn{12}{c}{\textbf{PERFORMANCE ON THE SIMULATOR DATASET}}\\
\hline
RetinaNet&ResNet-50 FPN&608$\times$800&36,276,717&\ding{55}&0.475&0.714&0.487&--&--&--&~25.0*\\
YOLOv3 & Darknet-53 &608$\times$800&61,576,342&\ding{55}&0.305&0.699&0.220&--&--&--&21.2\\
Poly-YOLO& SE-Darknet-53&608$\times$800&37,196,958&\ding{55}&0.413&0.735&0.408&--&--&--&22.0\\
Poly-YOLO& SE-Darknet-53 lite&416$\times$576&16,545,766&\ding{55}&0.322&0.661&0.258&--&--&--&58.6\\
\hline
Mask R-CNN&ResNet-50&448$\times$448&44,662,942&\ding{52}&0.389&0.664&0.414&0.203&0.452&0.157&15.8\\
Poly-YOLO& SE-Darknet-53& 608$\times$800&37,446,438&\ding{52}&0.435&0.745&0.445&0.345&0.731&0.272&19.6\\
Poly-YOLO& SE-Darknet-53 lite& 416$\times$576&16,712,302&\ding{52}&0.377&0.694&0.348&0.298&0.675&0.270&52.7\\
\hline
\hline
\multicolumn{12}{c}{\textbf{PERFORMANCE ON THE CITYSCAPES DATASET}}\\
\hline
RetinaNet&Resnet-50 FPN&608$\times$1216&36,504,912&\ding{55}&0.224&0.379&0.231&--&--&--&~21.0*\\
YOLOv3 & Darknet-53 &416$\times$832&61,640,962&\ding{55}&0.106&0.266&0.061&--&--&--&26.3\\
Poly-YOLO& SE-Darknet-53&416$\times$832&37,238,538&\ding{55}&0.168&0.344&0.141&--&--&--&26.5\\
Poly-YOLO& SE-Darknet-53 lite&320$\times$608&16,573,522&\ding{55}&0.104&0.231&0.080&--&--&--&46.8\\
\hline
Mask R-CNN&Resnet-50&1024$\times$1024&44,722,144&\ding{52}&0.164&0.318&0.151&0.069&0.202&0.031&6.2\\
Poly-YOLO& SE-Darknet-53&416$\times$832&37,488,018&\ding{52}&0.129&0.273&0.105&0.087&0.240&0.046&21.9\\
Poly-YOLO& SE-Darknet-53 lite&320$\times$608&16,740,058&\ding{52}&0.114&0.253&0.091&0.078&0.217&0.044&37.2\\
\hline
\hline
\multicolumn{12}{c}{\textbf{PERFORMANCE ON THE INDIA DRIVING DATASET}}\\
\hline
RetinaNet&Resnet-50 FPN&608$\times$1080&36,546,402&\ding{55}&0.221&0.357&0.230&--&--&--&~19.8*\\
YOLOv3 & Darknet-53 &448$\times$800&61,646,347&\ding{55}&0.117&0.267&0.089&--&--&--&23.9\\
Poly-YOLO& SE-Darknet-53&448$\times$800&37,242,003&\ding{55}&0.152&0.304&0.137&--&--&--&25.5\\
Poly-YOLO& SE-Darknet-53 lite&352$\times$608&16,575,835&\ding{55}&0.125& 0.260&0.105&--&--&--&46.7\\
\hline
Mask R-CNN&Renset-50&1024$\times$1024& 44,732,908 &\ding{52}&0.175&0.300&0.177&0.098&0.217&0.077&7.5\\
Poly-YOLO& SE-Darknet-53&448$\times$800&37,491,483&\ding{52}&0.145&0.288&0.134&0.115&0.267&0.083&20.6\\
Poly-YOLO& SE-Darknet-53 lite&352$\times$608&16,742,371&\ding{52}&0.131&0.263&0.119&0.101&0.239&0.074&37.1\\
\hline
\end{tabular}
\end{table*}

\subsection{Results}
\label{subsec-results}
We present all the measured results for the three datasets in Table~\ref{tab-all-results}. Poly-YOLO were trained separately for the version of pure bounding box detection and with bounding polygon detection. The original vanilla version\footnote{\url{github.com/qqwweee/keras-yolo3}} of YOLOv3, which we modified into the Poly-YOLO version is included as well. For the comparison with SOTA, we have trained RetinaNet\footnote{\url{github.com/facebookresearch/detectron2}} as a representative of the bounding box detection algorithms and Mask R-CNN\footnote{\url{github.com/matterport/Mask_RCNN}} as a representative of the instance segmentation algorithms. The SOTA algorithms were trained using transfer learning. Poly-YOLO was trained from scratch due to the fact that no pre-trained model is available. The models are trained until early stopping is reached. For the evaluation of results, we use the mAP coefficient from the official COCO repository\footnote{github.com/cocodataset/cocoapi/tree/master/PythonAPI}.

Overall, we can observe that Poly-YOLO significantly increases YOLOv3 detection accuracy (relative average increase is 40\%), although the inference speed is slightly faster. On the other hand, Poly-YOLO lite slightly outperforms the detection accuracy of YOLOv3, but it is twice faster. The important fact is also that Poly-YOLO with bounding polygon preserves the bounding box detection accuracy. Moreover, the accuracy of bounding boxes is increased in four out of six cases, so we can cay the features used for bounding polygons are suitable for bounding boxes too. If we compare Poly-YOLO with RetinaNet, we have to report that RetinaNet yields higher precision, but it is slower in two of three cases.

Let us also emphasize that the used framework and operating system have a significant impact. According to the original RetinaNet paper~\cite{lin2017focal}, it should run approximately 10FPS. But, the authors have rewritten the original RetinaNet implementation with the usage of the PyTorch framework, have optimized it massively and made it available only for Linux, which obviously leads to a significant speed-up. According to the official documentation of Detectron2\footnote{detectron2.readthedocs.io/notes/benchmarks.html} library from which we used RetinaNet implementation, the reimplementation increased computation speed three times. Therefore the comparison of the computation speed is slightly unfair for us as we are using Tensorflow and Windows. So, the open direction and the future work is to rewrite our Poly-YOLO to PyTorch a coding expert in order to reach even higher computation speed. Therefore, we mark RetinaNet inference speed in the table with * symbol. Let us note, YOLOv3 and Mask R-CNN were performed using the same framework and OS as Poly-YOLO, i.e., Tensorflow and Windows.

In the case of the simulator dataset where RetinaNet has the same resolution as Poly-YOLO, Poly-YOLO has higher AP$_{50}$, but it is less accurate for AP$_{75}$, which can be given by the fact that RetinaNet utilizes more anchor boxes, which can improve the precise detection. For the two other datasets, RetinaNet has been trained for a higher resolution (selected automatically), and it is better even for AP$_{50}$. When we analyzed the outputs, we observed that Poly-YOLO has a less precise class classification. That may be given by the fact that it uses a categorical cross-entropy for the classification loss function while RetinaNet uses focal loss that works significantly better for imbalanced datasets. Here, it may be beneficial to integrate the focal loss into Poly-YOLO in future work. From the last comparison with Mask R-CNN, we can report that Mask R-CNN has slightly better box detection accuracy, but it is less accurate in masking. Also, its processing speed is several times lower than the Poly-YOLO processing speed. That makes it useless for real-time image/video processing.

Note, the absolute values for all the neural networks is lower than the best possible values. That is by the fact our graphic cards are unable to process colossal batch sizes, and because of the training time, we do not realize the enormous amount of iterations with a fixed decrease of a learning rate, but we control the learning rate dynamically and utilize early stopping.

\section{Discussion}
\label{sec-ablation-study}
Here, we present the impact of hyperparameter setting on Poly-YOLO performance, discuss additional improvements and current limitations.

\subsection{Hyperparameters}
In Poly-YOLO, three aspects should be examined: the way, how squeeze-and-excitation blocks are integrated into the architecture, a dependency on a number of vertices, and a dependency on a number of anchors. 

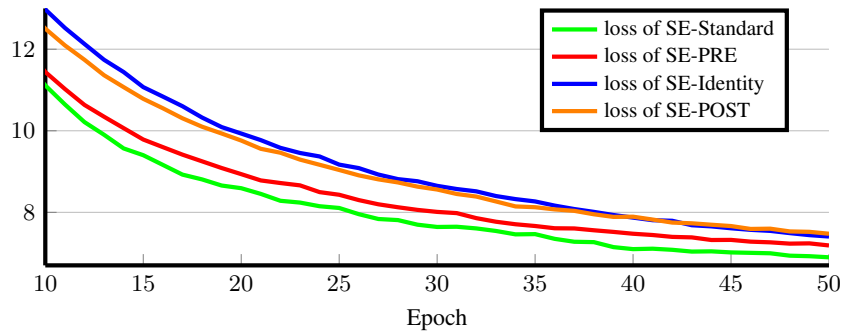
\begin{figure}[!h]
\centering
\begin{tikzpicture}
\small
\begin{axis}[height=50mm,width=120mm,xlabel={Epoch},xmin=10, xmax=50, ymin=6.7,ymax=13,legend style={at={(0.79,1)},anchor=north, nodes={scale=0.9, transform shape}},legend columns=1, ymajorgrids=true,yminorgrids=true,style={ultra thick},axis x line*=bottom, axis y line*=left, legend cell align={left}]
\addplot[green]coordinates{(1,269.1180581)(2,28.21265573)(3,18.40226598)(4,16.42302675)(5,15.21275809)(6,14.13723531)(7,13.14906785)(8,12.39609316)(9,11.69206872)(10,11.11319239)(11,10.64653439)(12,10.21312326)(13,9.90192655)(14,9.567035707)(15,9.401223558)(16,9.163649295)(17,8.925355853)(18,8.807712996)(19,8.657624128)(20,8.591970632)(21,8.452182244)(22,8.28339433)(23,8.239628017)(24,8.149788473)(25,8.107046533)(26,7.95557965)(27,7.836849302)(28,7.81011553)(29,7.697173212)(30,7.641633639)(31,7.646899995)(32,7.605880239)(33,7.543636947)(34,7.46002098)(35,7.466055063)(36,7.349977418)(37,7.27898359)(38,7.269490557)(39,7.147420347)(40,7.09822743)(41,7.108537794)(42,7.081504741)(43,7.037955232)(44,7.043813616)(45,7.017336998)(46,7.006936838)(47,6.996316055)(48,6.937507799)(49,6.924841831)(50,6.897511486)};\addlegendentry{loss of SE-Standard}
%\addplot[green,dashed]coordinates{(1,47.50182178)(2,30.68728327)(3,19.68424953)(4,18.29376094)(5,15.03333042)(6,15.33791863)(7,12.89367659)(8,12.2843216)(9,11.94397123)(10,11.56019573)(11,10.73157223)(12,10.5044645)(13,10.12754359)(14,10.38881053)(15,9.809634464)(16,9.612795427)(17,9.718599434)(18,9.197471745)(19,9.753666145)(20,9.072291718)(21,9.236122705)(22,8.766421022)(23,8.788898129)(24,8.958717487)(25,9.128627947)(26,8.551600418)(27,8.563221022)(28,8.688926815)(29,8.473807009)(30,8.366534389)(31,8.358534201)(32,8.182396103)(33,8.396462811)(34,8.271012892)(35,8.415828848)(36,8.167897335)(37,8.25534914)(38,8.181384558)(39,8.027670786)(40,8.060003098)(41,7.99971981)(42,7.990061356)(43,7.941530552)(44,7.85889572)(45,7.919655016)(46,7.905162687)(47,8.018190882)(48,7.85454203)(49,7.855816732)(50,7.799603132)};\addlegendentry{HC standard val loss}
\addplot[red]coordinates{(1,255.3146145)(2,27.95900442)(3,18.31181137)(4,16.39055273)(5,15.20779941)(6,14.19740698)(7,13.37207154)(8,12.64206936)(9,12.04604232)(10,11.44196109)(11,11.02513553)(12,10.63415717)(13,10.34436429)(14,10.06415156)(15,9.785490496)(16,9.599512872)(17,9.4151847)(18,9.25124012)(19,9.087143952)(20,8.936665181)(21,8.782586605)(22,8.718717793)(23,8.661535299)(24,8.496182178)(25,8.431155521)(26,8.300500657)(27,8.197329659)(28,8.124784009)(29,8.060078628)(30,8.010967848)(31,7.981127901)(32,7.857996452)(33,7.770133211)(34,7.708007798)(35,7.665074913)(36,7.609768813)(37,7.605236831)(38,7.560359192)(39,7.520371384)(40,7.475305292)(41,7.444159162)(42,7.398214444)(43,7.385829746)(44,7.321723127)(45,7.324325357)(46,7.281551455)(47,7.264726968)(48,7.2326907)(49,7.239353021)(50,7.188108735)
};\addlegendentry{loss of SE-PRE}
%\addplot[red,dashed]coordinates{(1,46.85067352)(2,41.90627698)(3,113.6842011)(4,66.81267617)(5,19.23694948)(6,14.2535718)(7,26.92453368)(8,12.5403826)(9,11.97699835)(10,11.64738649)(11,11.71448803)(12,11.08355352)(13,10.57902687)(14,10.29002868)(15,10.05079247)(16,10.43680592)(17,9.994612026)(18,9.745109104)(19,9.801051912)(20,9.52851997)(21,9.20885792)(22,9.296083986)(23,9.393848408)(24,9.31115346)(25,8.986179256)(26,9.020872484)(27,8.919367996)(28,8.789864496)(29,8.756568844)(30,8.890063324)(31,8.754691481)(32,8.614013309)(33,8.57510746)(34,8.486652939)(35,8.420117128)(36,8.354320625)(37,8.392999298)(38,8.374381575)(39,8.284394434)(40,8.393354391)(41,8.197058191)(42,8.397195877)(43,8.370758383)(44,8.14978912)(45,8.139622906)(46,8.124177055)(47,8.136123247)(48,8.113796421)(49,8.047958149)(50,8.019633734)};\addlegendentry{HC pre val loss}
\addplot[blue]coordinates{(1,264.8061946)(2,28.7521678)(3,18.82318241)(4,17.23244495)(5,16.17897792)(6,15.44333299)(7,14.74915674)(8,14.08886853)(9,13.48741767)(10,12.97578842)(11,12.52332146)(12,12.12831274)(13,11.74040393)(14,11.43977499)(15,11.07040893)(16,10.83877219)(17,10.60268841)(18,10.32508724)(19,10.09085365)(20,9.932103477)(21,9.770825149)(22,9.581551394)(23,9.455637853)(24,9.3704919)(25,9.170350541)(26,9.085107474)(27,8.926919662)(28,8.814437881)(29,8.762278944)(30,8.649891763)(31,8.571825632)(32,8.514846562)(33,8.399985339)(34,8.324478778)(35,8.267437839)(36,8.165272445)(37,8.080903234)(38,8.011588526)(39,7.934349906)(40,7.870824179)(41,7.807359677)(42,7.793145966)(43,7.683891885)(44,7.655667635)(45,7.609547111)(46,7.570819987)(47,7.545572116)(48,7.490602284)(49,7.441096358)(50,7.407082539)};\addlegendentry{loss of SE-Identity}
\addplot[orange]coordinates{(1,260.546094)(2,27.55126334)(3,18.68471213)(4,16.84290553)(5,15.76975086)(6,14.8353911)(7,14.13179376)(8,13.52979221)(9,13.00809115)(10,12.50434768)(11,12.09450333)(12,11.74126958)(13,11.36173982)(14,11.07000157)(15,10.7840254)(16,10.55237907)(17,10.30614998)(18,10.09999409)(19,9.933359726)(20,9.760630831)(21,9.557986649)(22,9.461992887)(23,9.293235519)(24,9.166584313)(25,9.039111848)(26,8.907204584)(27,8.808890323)(28,8.733564258)(29,8.630111975)(30,8.559798042)(31,8.450267491)(32,8.387168056)(33,8.267005898)(34,8.145155313)(35,8.125805241)(36,8.072470284)(37,8.036950649)(38,7.952647068)(39,7.888380086)(40,7.893282775)(41,7.823008168)(42,7.749308342)(43,7.731199059)(44,7.696206923)(45,7.6636727)(46,7.593042107)(47,7.599509333)(48,7.532048472)(49,7.522096642)(50,7.475368393)};\addlegendentry{loss of SE-POST}\end{axis}
\end{tikzpicture}
\caption{Progress of training loss for various SE block placing variants. The marked lines are computed as the mean from five runs.}
\label{fig-graph-se-type}
\end{figure}

According to the original paper~\cite{hu2018squeeze}, the best result is achieved when the squeeze-and-excitation block is placed in the first position in the residual block, before convolutions. However, Darknet-53 uses a special tuple of convolutions where the expansion layer follows the bottleneck layer. Therefore, we realized a simple experiment whose results are shown in Graph~\ref{fig-graph-se-type}. From this graph, we can see that SE-Standard, i.e., placing a squeeze-and-excitation block after the convolutions gives the best result. On the basis of this experiment, we use the setting of SE-Standard in our SE-Darknet-53.

\noindent

The second aspect is the maximal number of possible vertices in the polygon, or in other words, the resolution of the polar grid. Let us note, if an object is quadrangular, there is no difference if the maximal number is four or twenty. On the other hand, if an object is complicated and defined by, e.g., 80 vertices, with the maximal number set to 10, we will lose information due to quantization. In Graph~\ref{graph-nr-vertices}, we examine the impact of the maximal number of vertices on the loss and a mean average precision. According to the definition of the loss function given in Section~\ref{subsec-integration-with-yolo}, the error is summed over all the vertices. Therefore, a bigger maximal number of vertices should produce a higher loss if the objects consist of enough vertices. That assumption is reflected in the graph. It is interesting that the dependency is sub-linear. It means that increasing the maximal number of vertices does not produce a significantly more complicated task. On the contrary, increasing this number may lead to smaller quantization and deliver more information useful in the training of a network. The disadvantage is that as the number goes up, it increases the number of parameters in the high-resolution output tensor. That may force us to use smaller batch sizes and, therefore, to an increase in the training time. So, the proper selection is up to a user and available hardware.

The last aspect to investigate how precision depends on the number of anchors used. Let us recall, YOLOv2 uses a single output layer and five anchors, YOLOv3 uses three output layers with three anchors per layer, nine in total. When labels are pre-processed, each label is assigned to an anchor, for which the IOU is maximized. Then, such an anchor is used to detect a box for that particular label the network is trained. From that, it is evident that the higher number of anchors makes the task more complicated. On the other hand, a higher number of anchors may be helpful to partially solve the label rewriting problem mentioned in Section~\ref{subsec-yolov3-issues}. In Graph~\ref{graph-nr-vertices}, we show results of the experiment where a network is trained for a various number of anchors. According to the loss, the optimal value is between six and nine anchors; a higher or lower number increases both the training and the validation loss. We have selected the same number as YOLOv3, i.e., nine.

\subsection{Emphasizing parts of detections}
\begin{figure}[!h]
    \centering
    \includegraphics[width=82mm]{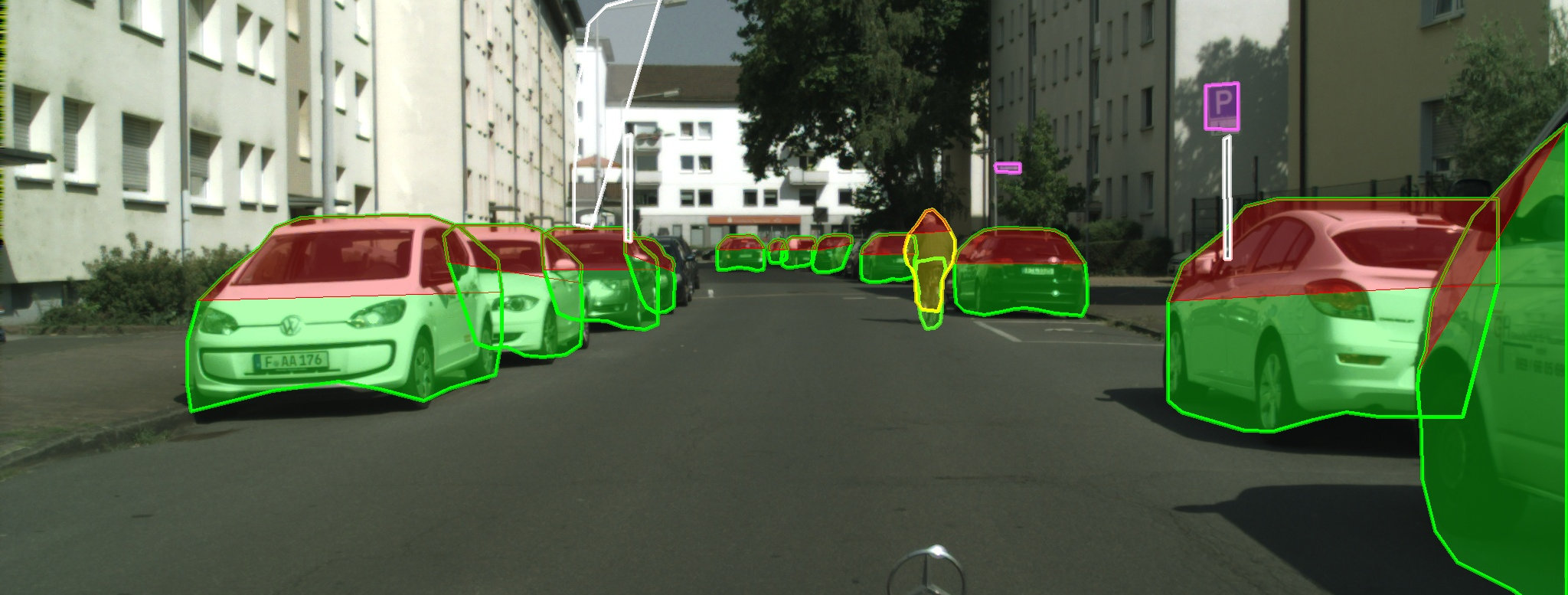}
    \includegraphics[width=82mm]{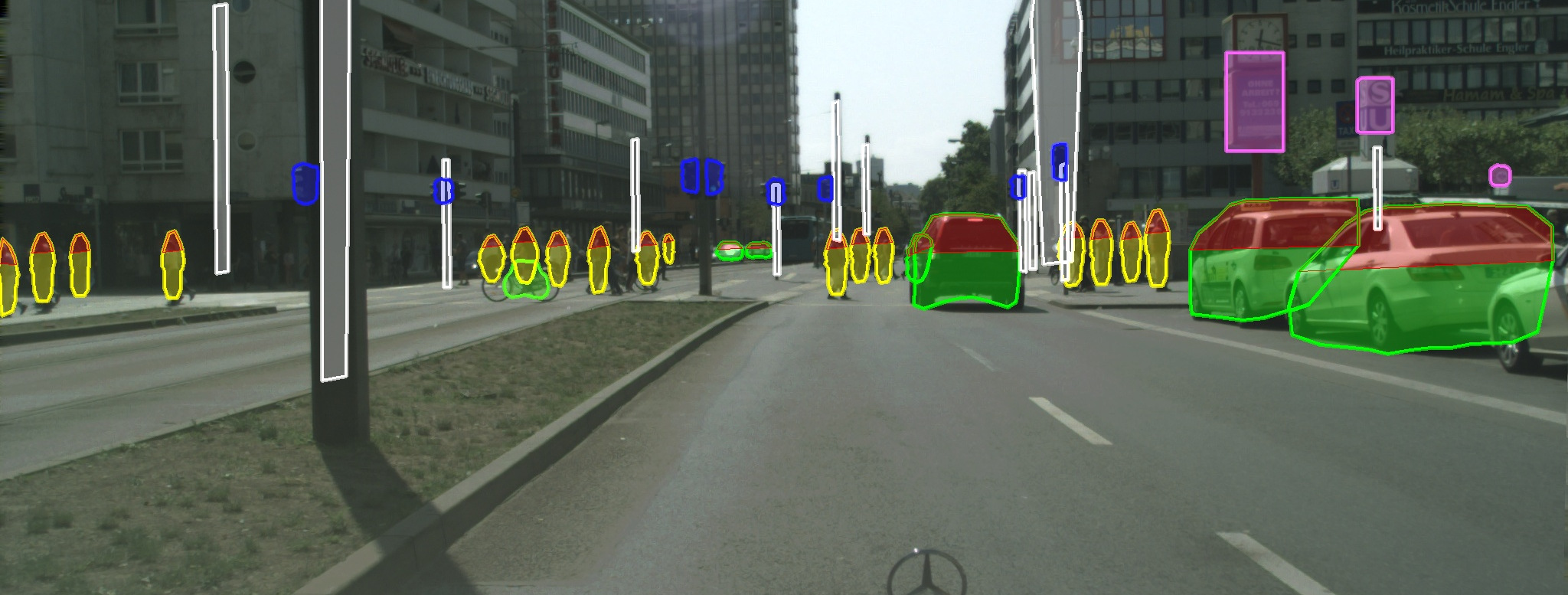}
    \caption{The two images illustrate the case when there are manually defined intervals for angles of vertices (for cars and for pedestrians). Object area defined by these vertices should be dimmed while the rest of the object should be emphasized by car headlamps.}
    \label{fig-glass-head}
\end{figure}

As we mentioned in Section~\ref{sec-problem-statement}, the practical application where quick instance segmentation may be helpful is the implementation of an intelligent car headlamp, where various objects in front of a car can be lightened/dimmed individually. The precise object detection based on the polygonal principle, which we propose, uses polar coordinates that allow improving the functionality even further. Let us consider classes such as a car, a biker, a pedestrian, or a van. Objects with such a class should be lightened more (to increase their visibility), but it is also necessary to avoid their dazzling. The solution is not to illuminate parts that include a front glass of a car, head of a pedestrian, etc. To detect such parts, additional extensive data labeling would be required in the standard case. In our case, it is enough to manually define an interval in the polar coordinate system that should not be dazzled. The benefits are that such manual definition is fast, easily controlled, it does not affect training/inference speed, it is explainable, and what is essential -- it is independent on a size of an object or its aspect ratio. Finally, it is not necessary to define additional labels. The illustration of such inference with this additional extension is shown in Figure~\ref{fig-glass-head}.

\subsection{Limitations}
\begin{figure}[!h]
    \centering
    \includegraphics[height=45mm]{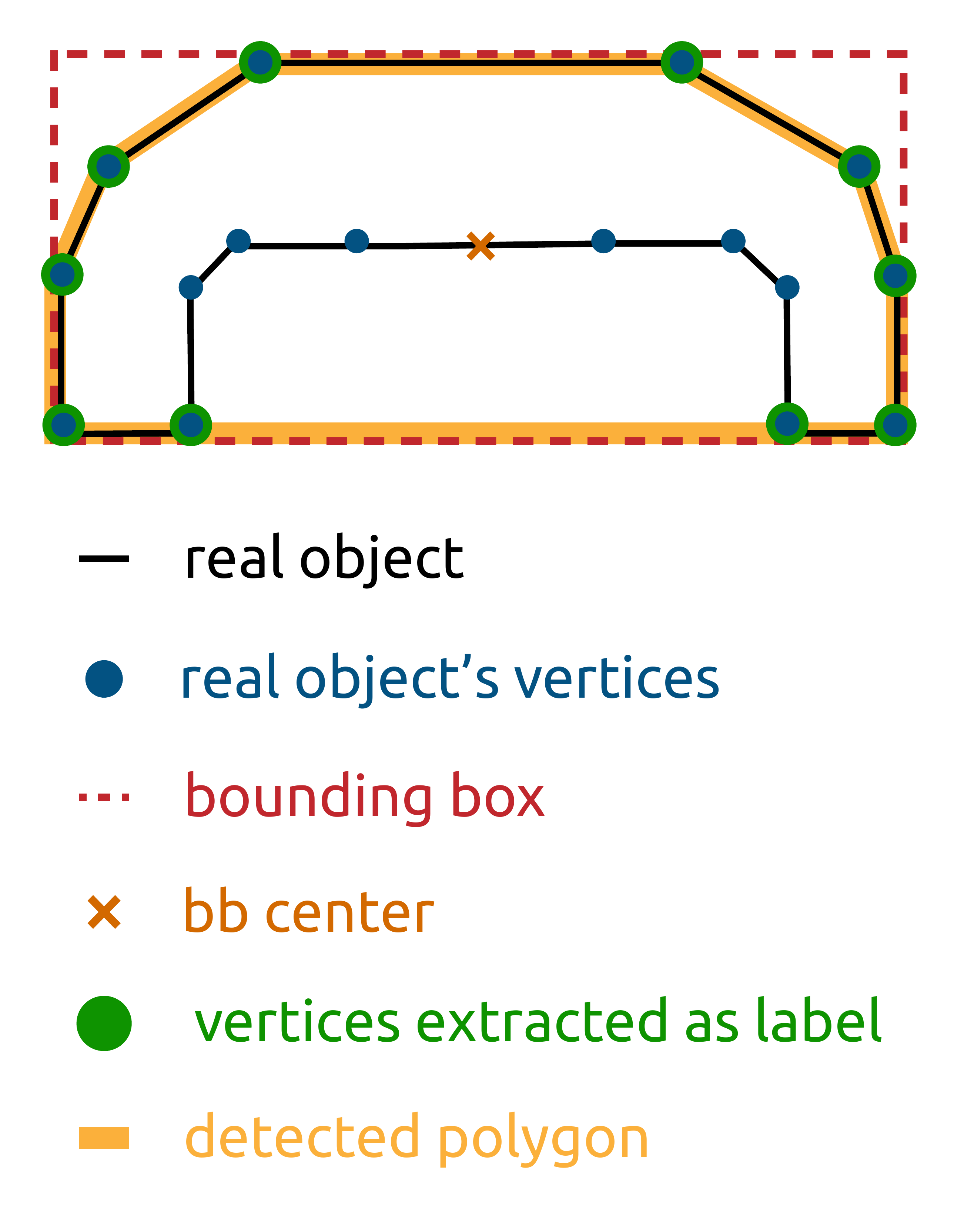}
    \includegraphics[height=45mm]{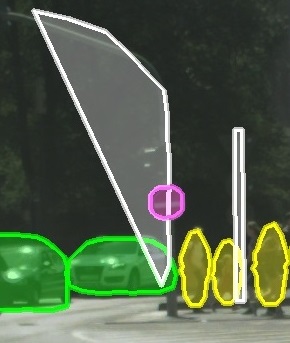}
    \caption{Left:A scheme of label creation for a problematic object, where the limitation appear. Right: impact on a real predictions, see the lamp object.}
    \label{fig-limitations-convex}
\end{figure}

The major limitation discovered during our research came from the scheme used for polygon vertices definition and the scheme of how the labels are created, as is described in Section~\ref{ref-subsec-preparing-data}. If two vertices belong to the same polar cell, the vertex with a bigger distance from the bounding box center is taken. That may lead to a situation when a new part to a strongly non-convex object is added, as it is shown in Figure~\ref{fig-limitations-convex}.  The figure also shows the practical impact of this limitation.  Let us note, it is not a problem of training or inference. It is a problem of the creation of labels; the network itself is trained correctly and makes predictions based on the (imprecise) training labels. After we connect individual vertices, we connect the first vertex with the last one, and this can cause problems for strongly non-convex objects. For completeness, this behavior does not happen for all non-convex objects. If two vertices lie in two distinct polar cells, even non-convex objects will be handled correctly, as can be seen in Figure~\ref{fig-precision-of-bounds}, where Poly-YOLO works nicely even for non-convex stars.

\section{Summary}
We have presented Poly-YOLO, which improves YOLOv3 in three aspects. It is more precise, faster, and able to realize instance segmentation. The precision is improved due to the analysis of issues in YOLO (rewriting of labels and incorrect distribution of anchors) and their removal by a newly proposed neck and head. The neck consists of the hypercolumn technique improved by the stairstep approach, and the head processes a single output tensor with high resolution. The new neck and head reach higher precision, which allows us to decrease the number of parameters in the original feature extractors, while still preserving a significantly higher precision. Poly-YOLO has only 60\% of parameters of YOLOv3 but improves the accuracy by relative 40\%. 

For the task of instance segmentation, we have designed an extension that detects bounding polygons with a dynamic number of vertices per detected object. The proposed bounding polygon detection learns itself size-independent shapes, which simplifies the task.  Poly-YOLO is able to run real-time on mid-tier graphics cards.

\section*{Acknowledgment}
The work is supported by ERDF/ESF "Centre for the development of Artificial Intelligence Methods for the Automotive Industry of the region" (No. CZ.02.1.01/0.0/0.0/17049/0008414)

\bibliographystyle{unsrt}  
\bibliography{literature}  %%% Remove comment to use the external .bib file (using bibtex).

\end{document}